\documentclass[10pt,conference,letterpaper]{IEEEtran}
\usepackage{booktabs} 
\usepackage{amsmath,url,graphicx,amssymb}
\usepackage{amsfonts}
\usepackage{ctable}
\usepackage{multirow}
\usepackage{algorithm}
\usepackage{algpseudocode}
\usepackage{pifont}
\usepackage{color}
\usepackage{subfigure}
\usepackage{sidecap}

\newcommand{\mylistbegin}{
  \begin{list}{$\bullet$}
   {
     \setlength{\itemsep}{-2pt}
     \setlength{\leftmargin}{1em}
     \setlength{\labelwidth}{1em}
     \setlength{\labelsep}{0.5em} } }
\newcommand{\mylistend}{
   \end{list}  }

\newcommand{\eg}{\textit{e.g.}}

\newcommand{\ie}{\textit{i.e.}}

\newcommand{\wrt}{\textit{w.r.t.~}}
\newcommand{\header}[1]{{\vspace{+1mm}\flushleft \textbf{#1}}}

\newtheorem{theorem}{Theorem}[section]

\newtheorem{definition}[theorem]{Definition}

\title{Meta-Graph Based HIN Spectral Embedding: Methods, Analyses, and Insights}
\author{
{Carl Yang, Yichen Feng, Pan Li, Yu Shi, Jiawei Han}
\vspace{1.6mm}\\
\fontsize{10}{10}\selectfont\itshape
University of Illinois, Urbana Champaign, 201 N Goodwin Ave, Urbana, Illinois 61801, USA\\
\fontsize{9}{9}\selectfont\ttfamily\upshape
\{jiyang3, feng36, panli2, yushi2, hanj\}@illinois.edu
}

\begin{document}

\setlength{\floatsep}{4pt plus 4pt minus 1pt}
\setlength{\textfloatsep}{4pt plus 2pt minus 2pt}
\setlength{\intextsep}{4pt plus 2pt minus 2pt}
\setlength{\dbltextfloatsep}{3pt plus 2pt minus 1pt}
\setlength{\dblfloatsep}{3pt plus 2pt minus 1pt} 
\setlength{\abovecaptionskip}{3pt}
\setlength{\belowcaptionskip}{2pt}
\setlength{\abovedisplayskip}{2pt plus 1pt minus 1pt}
\setlength{\belowdisplayskip}{2pt plus 1pt minus 1pt}

\maketitle
%!TEX root = hinse.tex
\begin{abstract}
Heterogeneous information network (HIN) has drawn significant research attention recently, due to its power of modeling multi-typed multi-relational data and facilitating various downstream applications. 
In this decade, many algorithms have been developed for HIN modeling, including traditional similarity measures and recent embedding techniques. Most algorithms on HIN leverage \textit{meta-graphs} or \textit{meta-paths} (special cases of meta-graphs) to capture various semantics. 
Given any arbitrary set of meta-graphs, existing algorithms either consider them as equally important or study their different importance through supervised learning. Their performance largely relies on prior knowledge and labeled data.
While unsupervised embedding has shown to be a fundamental solution for various homogeneous network mining tasks, for HIN, it is a much harder problem due to such a presence of various meta-graphs.

In this work, we propose to study the utility of different meta-graphs, as well as how to simultaneously leverage multiple meta-graphs for HIN embedding in an unsupervised manner. Motivated by prolific research on homogeneous networks, especially spectral graph theory, we firstly conduct a systematic empirical study on the spectrum and embedding quality of different meta-graphs on multiple HINs, which leads to an efficient method of meta-graph assessment. It also helps us to gain valuable insight into the higher-order organization of HINs and indicates a practical way of selecting useful embedding dimensions.
Further, we explore the challenges of combining multiple meta-graphs to capture the multi-dimensional semantics in HIN through reasoning from mathematical geometry and arrive at an embedding compression method of autoencoder with $\ell_{2,1}$-loss, which finds the most informative meta-graphs and embeddings in an end-to-end unsupervised manner.
Finally, empirical analysis suggests a unified workflow to close the gap between our meta-graph assessment and combination methods.
To the best of our knowledge, this is the first research effort to provide rich theoretical and empirical analyses on the utility of meta-graphs and their combinations, especially regarding HIN embedding.
Extensive experimental comparisons with various state-of-the-art neural network based embedding methods on multiple real-world HINs demonstrate the effectiveness and efficiency of our framework in finding useful meta-graphs and generating high-quality HIN embeddings.
\end{abstract}

%\keywords{heterogeneous information networks, network embedding, unsupervised meta-graph selection}

%!TEX root = hinse.tex
\section{Introduction}
\label{sec:intro}
Networks are widely used to model relational data such as web pages with hyperlinks and people with social connections. Recently, increasing research attention has been paid to the heterogeneous information network (HIN), due to its power of accommodating rich semantics in terms of multi-typed nodes (vertices) and links (edges), which enables the integration of real-world data from various sources and facilitates wide downstream applications \cite{sun2011pathsim,liu2017semantic,hou2017hindroid,sun2013pathselclus,zhao2017meta,yang2018did}.

For capturing the complex semantics in HIN, the concepts of meta-paths and meta-graphs have been developed, which are subsets of HIN schemas \cite{sun2012mining}. Since each particular meta-graph indicates an essential semantic unit that can be potentially useful for various tasks, they have become the de facto tool of HIN modeling, leveraged by various existing works \cite{sun2011pathsim,wang2016relsim,fang2016semantic,meng2015discovering,shi2017prep,wang2015knowsim,dong2017metapath2vec,shang2016meta,huang2017heterogeneous,fu2017hin2vec,shi2018aspem,wan2015graph}. In this work, to be general, we refer meta-paths to special cases of meta-graphs, and study them under the same framework. 

Since there can be various meta-graphs on a given HIN, the key problems for leveraging them are: (1) what meta-graphs are useful (\textit{assessment}), and (2) how to jointly leverage multiple meta-graphs (\textit{combination}). To our surprise, however, no existing work explicitly studies the first problem, while no satisfactory solution exists to the second, especially regarding general-purpose unsupervised network embedding.

To get around the first problem, existing HIN models mostly assume that useful meta-graphs can be manually composed based on domain knowledge \cite{sun2011pathsim,wang2016relsim,shi2017prep,wang2015knowsim,dong2017metapath2vec,shang2016meta,huang2017heterogeneous,fu2017hin2vec,wan2015graph,hou2017hindroid,sun2013pathselclus}, while such knowledge can be expensive and not always available for arbitrary unfamiliar HINs. To break this limitation, a few algorithms attempt to generate all legitimate meta-graphs up to a certain size through heuristic mechanisms \cite{fang2016semantic,meng2015discovering,shi2018aspem,jiang2017semi}, but they again fail to further select the more useful ones before sending all of them into a subsequent combination model.

As for the second problem, most algorithms rely on supervised learning towards specific tasks to tune the weights on different meta-graphs \cite{wang2016relsim,hou2017hindroid,chen2016task,yu2012user,zhao2017meta,meng2015discovering,shi2017prep}. Their performance heavily relies on labeled data. On the other hand, while general-purpose unsupervised network embedding has received tremendous attention recently due to the huge success of neural network based models like \cite{perozzi2014deepwalk,tang2015line,grover2016node2vec,tang2015pte}, there exists no method to properly combine multiple meta-graphs for unsupervised HIN embedding, except for simply adding up their instance counts \cite{dong2017metapath2vec,huang2017heterogeneous} or looking for the proper weights through exhaustive grid search \cite{shang2016meta}.
%hin2vec

%To leverage meta-graphs for various downstream tasks, existing HIN models either compute a similarity score based on the number of meta-graph instances between two nodes \cite{sun2011pathsim,wang2016relsim,fang2016semantic,meng2015discovering,shi2017prep,wang2015knowsim}, or learn a network embedding based on sampled instances of meta-graphs \cite{dong2017metapath2vec,shang2016meta,huang2017heterogeneous,fu2017hin2vec,shi2018aspem,wan2015graph}. Without precise domain knowledge, existing methods often enumerate a large set of meta-graphs and then select the useful ones based on supervised learning \cite{hou2017hindroid,sun2013pathselclus,fang2016semantic,meng2015discovering,wang2016relsim}. machines through exhaustive enumeration, which fall short when facing real-world unfamiliar complex networks. Moreover, when considering multiple meta-graphs, most algorithms simply add up their instance counts and compute a single model in one common space \cite{dong2017metapath2vec,shang2016meta,huang2017heterogeneous,sun2011pathsim,wang2016relsim,fang2016semantic,meng2015discovering,wang2015knowsim}, which ignores the geometric property of the underlying graphs induced by different meta-graphs. Some of them are able to learn different importance weights for the considered meta-graphs, but they all do it in a supervised way, which heavily relies on labeled data \cite{wang2016relsim,hou2017hindroid,chen2016task,yu2012user,zhao2017meta,meng2015discovering,shi2017prep}.

In this work, we extend the rich theoretical and empirical studies on homogeneous networks to the HIN setting. Specifically, we provide a series of \textit{methods}, \textit{analyses} and \textit{insights} towards \textit{meta-graph based HIN spectral embedding}, which serves as solutions to both of the aforementioned assessment and combination problems. 
Our main contributions are summarized in the following.

\header{Contribution 1: Meta-Graph based HIN Spectral Embedding.}
Motivated by prolific studies on homogeneous networks, we review and introduce several key conclusions from spectral graph theory, and propose to leverage meta-graphs to compute the \textit{projected networks} of HIN. It facilitates HIN spectral embedding, which serves as a great tool for various subsequent theoretical and empirical analyses (Section \ref{sec:related} and \ref{sec:embedding}).

\header{Contribution 2: Meta-Graph Assessment.}
Based on well-established \textit{spectral graph theory}, we compute the graph spectra of projected networks, which in principle capture the key network properties. 
Through a systematic empirical study on three real-world HINs, we discover two essential properties that have significant impacts on the general quality of HIN embedding. Theoretical interpretations of these properties provide valuable insights into the high-order organizations of HINs and their implications towards embedding quality, which further allows efficient assessment of meta-graph utility (Section \ref{sec:assessment}).  

\header{Contribution 3: Meta-Graph Combination.}
Since different meta-graphs essentially capture different semantic information of a HIN, it is necessary to properly combine multiple useful meta-graphs. 
To simultaneously solve the intrinsic \textit{dimension reduction} and \textit{meta-graph selection} problems in an \textit{unsupervised} manner, we devise an autoencoder with $\ell_{2,1}$-loss. It is able to end-to-end select the important meta-graphs from a set of candidates by capturing the embedding dimensions with large variance grouped by the corresponding meta-graphs. We also provide rich theoretical and empirical analyses towards its effectiveness (Section \ref{sec:combination}).
%Specifically, for HIN embedding, concatenating multiple embeddings computed from individual meta-graphs ignores the interactions and correlations among meta-graphs, and often leads to redundant information.
%On the other hand, from the mathematical geometry point of view, embedding multiple meta-graphs into a common space corresponds to a distortion to their individual spectral embedding, which always incurs an information loss. To find a generally useful embedding, we need to put more stress on the more informative meta-graphs. 
\header{Contribution 4: Comprehensive Evaluations.}
Through extensive experiments in comparison with various state-of-the-art HIN embedding methods on three large real-world datasets towards two traditional downstream tasks, we demonstrate the supreme performance of our proposed method for general-purpose unsupervised HIN embedding (Section \ref{sec:exp}).

%!TEX root = hinse.tex
\section{Related Work and Preliminaries}
\label{sec:related}
\subsection{Heterogeneous Information Network (HIN)}
Networks provide a natural and generic way of modeling data with interactions. Among them, HIN has drawn increasing research attention in the recent decade, due to its capability of retaining rich type information \cite{sun2012mining}.
A HIN can be defined as $\mathcal{N}_H = (\mathcal{V}_H, \mathcal{E}_H, \phi, \psi)$, where $\mathcal{V}_H$ is the vertex set and $\mathcal{E}_H$ is the edge set.
In addition to traditional homogeneous networks, $\phi:\mathcal{V}_H\rightarrow \mathcal{T}=(t_1, t_2,...)$ and $\psi: \mathcal{E}_H\rightarrow \mathcal{R}=(r_1, r_2,...)$ are two mapping functions that assign vertices and edges with the type information.
Such extensions, while making the networks much more complicated, have shown to be very powerful in modeling real-world multi-modal multi-aspect data \cite{sun2011pathsim,wang2016relsim,meng2015discovering,jiang2017semi,wang2015knowsim} and beneficial to various downstream applications \cite{liu2017semantic,hou2017hindroid,sun2013pathselclus,zhao2017meta,yang2018did}.
To model HIN with typed vertices and edges, \cite{sun2011pathsim} proposes to leverage the tool of meta-paths, which is later on generalized to meta-graphs \cite{fang2016semantic}. They are adopted by almost all HIN models due to the capture of fine-grained type-and-structure-aware semantics.

Recently, network embedding algorithms based on the advances in neural networks (NN) have been extremely popular \cite{perozzi2014deepwalk,tang2015line,grover2016node2vec,tang2015pte}. 
They aim to compute distributed representations of vertices that capture both neighborhood and structural similarity \cite{wang2016structural,lyu2017enhancing}. 
%The learned embedding vectors can then be leveraged for various network learning tasks including link prediction \cite{zhang2017weisfeiler}, node classification \cite{yang2016revisiting}, community detection \cite{yang2017graph}, \etc. 
Following this trend, many HIN embedding methods have also been developed \cite{dong2017metapath2vec,shang2016meta,huang2017heterogeneous,fu2017hin2vec,shi2018aspem,wan2015graph,shi2017prep}. Most of them, while guided by meta-graphs, mainly leverage well-developed NN models (\eg, Skip-gram \cite{mikolov2013distributed}). 
While they are shown to work well in certain cases, their performances are not stable and hard to track.  

In this work, we get inspired by prolific studies on homogeneous networks. For the first time, we provide a series of theoretically sound and empirically effective methods towards HIN embedding, together with extensive analyses and valuable insights, based on the well-established spectral graph theory.

\subsection{Spectral Graph Theory}
%Spectral graph theory has a long history \cite{chung1997spectral}. In the early days, matrix theory and linear algebra were used to analyze the adjacency matrices of graphs, which are especially effective for regular and symmetric graphs. 
Spectral embedding, also termed as the Laplacian eigenmap, has been widely used for homogeneous network embedding \cite{belkin2003laplacian,white2005spectral}. 
Mathematically, it can be computed as follows: Given a weighted homogeneous network $\mathcal{G}=(\mathcal{V}, \mathcal{E})$, where $\mathcal{V}$ is the vertex set and $\mathcal{E}$ is the edge set. We also define an adjacency matrix $\mathcal{A}$. For any $e_{ij} \in \mathcal{E}$, $a_{ij}>0$ denotes its edge weight, and for any $e_{ij} \notin \mathcal{E}$, $a_{ij}=0$. Let $\mathcal{D}$ be the diagonal degree matrix where $d_{ii} = \sum_{e_{ij}\in \mathcal{E}} a_{ij}$. Then, the \textit{normalized Laplacian matrix} follows 
\begin{align}
\mathcal{L} = I - \mathcal{D}^{-\frac{1}{2}} \mathcal{A} \mathcal{D}^{-\frac{1}{2}},
\label{eq:lap}
\end{align}
where $I$ is the identity matrix. Suppose $\mathcal{L}$ has eigenvalues ordered as $\lambda_1=0 \leq \lambda_2 \leq \cdots \leq \lambda_{|\mathcal{V}|}$, which is also termed as the spectrum of graph $\mathcal{G}$. For each eigenvalue $\lambda_i$, we denote the corresponding eigenvector as $u_i = (u_{ij})_{v_j\in \mathcal{V}}$. Then, the $k$-dimensional embedding of vertice $v_j$ can be expressed as $\mathbf{h}_j = [u_{1j}, u_{2j},...,u_{kj}]^T$.  
Spectral graph theory connects the spectrum of $\mathcal{L}$ to the properties of $\mathcal{G}$ and further gives plentiful results that are useful in both theory and practice. 

For later usage, in the following, we review some key theories and definitions related to our work while refer interested readers to \cite{chung1997spectral,lee2014multiway,li2018submodular} for more results. 

\begin{theorem}[\cite{chung1997spectral}] \label{thm:conn}
The number of zero eigenvalues of $\mathcal{L}$ is equal to the number of connected components of $\mathcal{G}$. 
\end{theorem}

Suppose the number of zero eigenvalues is $p$. One step further, the first $p$ dimensions of $h_i$ are orthogonal to those of $h_{j}$ if $v_i$ and $v_j$ lie in different connected components. So spectral embedding naturally encodes the connectivity between vertices in the embedding space. 

We next introduce some results on the concept of \textit{nodal domain} \cite{davies2001discrete,tudisco2016nodal} that may be used to understand the embedding space. We start with the definition of some key concepts. 
\begin{definition}
For a subset of vertices $\mathcal{S}\subseteq \mathcal{V}$, we denote the \textit{induced subgraph} of $\mathcal{G}$ by $\mathcal{S}$ as $\mathcal{G}(\mathcal{S}) = (\mathcal{S}, \mathcal{E}(\mathcal{S}))$, such that for any pair of vertices $v_i, v_j\in \mathcal{S}$, the edge $e_{ij} \in \mathcal{E}(\mathcal{S})$ if and only if $e_{ij}\in \mathcal{E}$.  
\end{definition}
\begin{definition}
Given a function $f(\cdot): \mathcal{V}\rightarrow \mathbb{R}^{|\mathcal{V}|}$. A subset $\mathcal{S}\subset \mathcal{V}$ is a \textit{strong nodal domain} of $\mathcal{G}$ induced by $f$ if the induced subgraph $\mathcal{G}(\mathcal{S})$ is a maximal connected component of either $\{v_i\in \mathcal{V} : f(v_i) > 0\}$ or $\{v_i \in \mathcal{V} : f(v_i) < 0\}$.
\end{definition} 

Next we introduce two powerful results from spectral graph theory that characterize the nodal domains of an eigenvector. 

The first one gives the bound on the number of nodal domains of an eigenvector.
\begin{theorem}[\cite{davies2001discrete}] \label{thm:nodal-domain}
If the network is connected, the number of strong nodal domains of $v_i$ is not greater than the number of eigenvalues that are not greater than $\lambda_i$.
\end{theorem}

It implies that for small $\lambda_i$, the number of nodal domains induced by $v_i$ is also small.

The next one is the high-order Cheeger inequality. 
\begin{theorem}[\cite{tudisco2016nodal}] \label{thm:high-order-cheeger}
Suppose $\{\mathcal{S}_l\}_{1 \leq l\leq m}$ are strong nodal domains induced by $v_i$. Then 
\begin{equation}
\max_{1\leq l\leq m} \frac{\sum_{j\in \mathcal{S}_l, j'\notin \mathcal{S}_l} a_{jj'}}{\sum_{j\in \mathcal{S}_l, j'\in V} a_{jj'}} \leq \sqrt{2 \lambda_i}. 
\label{eq:high-order-cheeger}
\end{equation}
\end{theorem}

It implies that each nodal domain associated with small eigenvalues corresponds to the community structures of $\mathcal{G}$, whose inside is densely connected with few out-edges. 

These two theorems indicate how spectral embedding represents the topology of $\mathcal{G}$ in the Euclidean space. 
As we will see in the remainder of this work, these theories lay the solid foundation of our methods, guide our fruitful data analyses and lead to quite a few valuable insights.

%
%\begin{itemize}
%
%\item Small eigenvalues characterize the clustering tendency of nodes, which is established by high-order cheeger inequalities. 
%
%\end{itemize}
%
%
%
%that yields spectral embedding
%
%Spectral embedding is mathematically equivalent to 
%
%Cheeger inequalities tightly bound the distortion induced by the low-dimensional embedding with respect to ... norm. Moreover, nodal domain theorems explain the structure of graphs via harmonic analysis. figure. One of our critical findings built upon spectral graph theory shows that the mostly useful embedding should be the segment of Laplacian eigenmap that come from the low-frequent harmonic structure  within each connected components of the obtained graph after meta-graph aggregation. However, we find the identity of each component are not informative, which is in contrast to those observed in homogeneous information network.

%!TEX root = hinse.tex
\section{Meta-Graph Based HIN Spectral Embedding}
\label{sec:embedding}
In order to generalize the key theoretical studies and empirical analyses on homogeneous networks to HIN, we introduce our basic HIN spectral embedding method.

Traditional graph theory studies the adjacency matrices of homogeneous networks.
As we discussed in Section \ref{sec:related}.A, the additional type information endows HIN with advantageous modeling capability but also makes it much more complicated and inappropriate to be represented by a single adjacency matrix.
To this end, we leverage the powerful tool of meta-graphs that encode various fine-grained HIN semantics by designing a HIN projection process. 
While spectral embedding has been widely studied \cite{yin2017local,zhou2017local,long2006spectral,sengupta2015spectral}, no previous work has connected it with the utility of meta-graphs on HIN.

\begin{figure}[h!]
    \centering
        \includegraphics[width=0.9\linewidth]{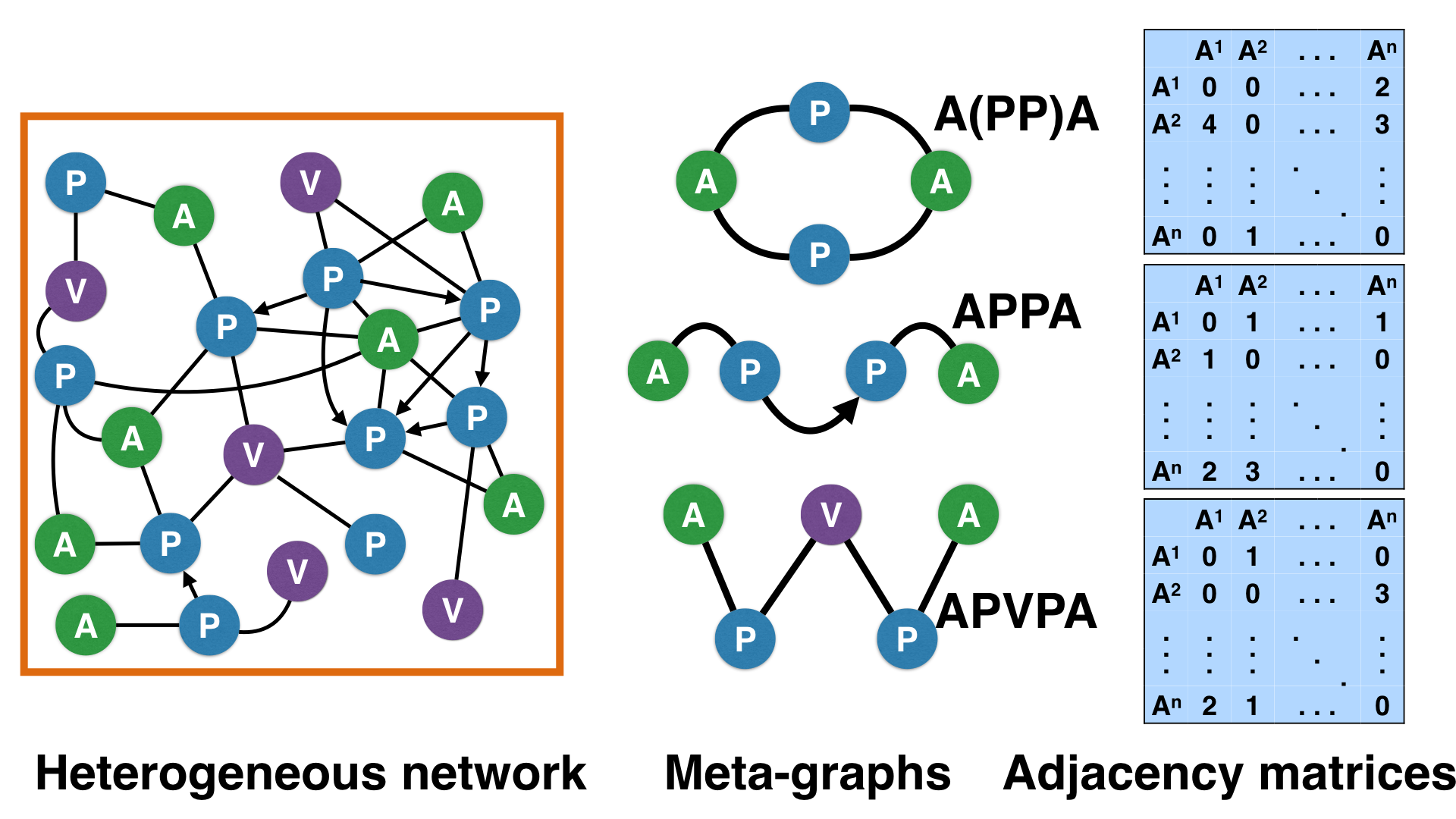}
        \vspace{-10pt}
    \caption{The process of obtaining the projected homogeneous networks and corresponding adjacency matrices from HINs with different meta-graphs.}
    \vspace{-5pt}
    \label{fig:project}
 \end{figure}

Figure \ref{fig:project} shows an example of the academic publication network, where we use three different meta-graphs to project the HIN and get three different adjacency matrices for the corresponding homogeneous networks of authors. The edge weights are generated by the number of matched meta-graph instances between each pair of vertices. We call the homogeneous networks obtained in this way the \textit{projected networks}. 
Note that, during this procedure, the type information is captured by meta-graphs which may further be encoded into the edge-weights of the projected networks. Therefore, one may expect to obtain a good vertex embedding as long as the meta-graphs are chosen properly. 

In Figure \ref{fig:project}, we also give a few examples of meta-graphs and their notations. For simplicity, we only consider edges connecting to the pairs of vertices (\eg, authors on the two sides here), and do not differentiate directed and undirected edges, while our methods trivially generalize to those cases.

Based on the homogeneous projected networks, we can compute the standard spectral embedding as described in Section \ref{sec:related}.B. 
Note that spectral embedding is mathematically equivalent to the PCA of a degree normalized adjacency matrix $\mathcal{A} = I + D^{-\frac{1}{2}} A D^{-\frac{1}{2}}$ \cite{wold1987principal}, so approximating the original graph in the optimal sense such that $H=(h_1^T, h_2^T, ..., h_{|\mathcal{V}|}^T)^T$ gives the solution to the approximation problem 
\begin{align}\label{eq:pca}
\min_{H\in\mathcal{R}^{|\mathcal{V}|\times k}, W\in \mathbb{R}^{k\times k}}\| \mathcal{A} -  HWH^T\|_{F}^2,
\end{align} 
where $W$ is a diagonal matrix and $\|\cdot\|_{F}$ is Frobenius norm.

In contrast to those complex NN-based approaches, spectral embedding holds superiority in several aspects. 
First, it is computationally cheaper. For a $k$-dimensional embedding, one may require $O(k|\mathcal{E}|)$ number of scalar sum and product operations on the projected networks based on power iterations \cite{davidson1975iterative}, while one single epoch of training the NN-based models costs such amount of computation on the original HIN with much more vertices and edges. 
More importantly, unlike the NN-based approaches that implicitly factorize the adjacency matrices \cite{qiu2018network}, spectral embedding directly provides linear approximation for the PCA problem in Eq.~\ref{eq:pca}. Its performance is more tractable through the well-established spectral graph theory, which makes it a good tool to understand the underlying structures and principal properties of networks, as well as the function of different meta-graphs.

In spectral embedding, an eigenvector $u_i$ can be viewed as a one-dimensional embedding of vertices. Conceptually, based on Theorem \ref{thm:nodal-domain} and \ref{thm:high-order-cheeger}, for a small eigenvalue $\lambda_i$, the vertices from one nodal domain of $v_i$ typically lie within a densely connected community of $\mathcal{G}$. Correspondingly, due to the definition of nodal domains, all the vertices in this nodal domain will be embedded into the same quadrant in the embedding space. This relation gives a direct mapping from the densely connected communities of $\mathcal{G}$ to a quadrant of the embedding space. As each of the eigenvectors can be viewed as a one-dimensional embedding as described above, the spectral embedding based on the concatenation of eigenvectors with small $\lambda_i$ actually gives a fine embedding of the whole graph $\mathcal{G}$ in the sense that vertices topologically close on $\mathcal{G}$ are essentially more likely to be embedded into the same quadrant. Moreover, it also makes the change and tuning of embedding sizes extremely efficient. To increase the embedding size by $k'$, only $O(k'|\mathcal{E}|)$ time is required, while decreasing the embedding size takes no time. On the other hand, the NN-based models need to be totally retrained whenever the embedding sizes are changed.

Our method is also closely related to the spectral methods leveraged for the investigation of higher-order organizations of homogeneous networks in \cite{benson2016higher,yin2017local}. 
However, in HIN, the high-order connectivity patterns are carried by meta-graphs that encode various semantic information. Moreover, the projection process is quite different, since meta-graphs do not always lead to cliques as the network motifs in \cite{benson2016higher}. A very recent work on hypergraphs shows that the spectral clustering based on inhomogeneous projections of \textit{hyperedges} keeps good approximation of the cheeger isoperimetric constant of hypergraphs~\cite{li2017inhomogeneous}. Since hyperedges can be viewed as a mathematical abstract of our meta-graphs, this implies that our method essentially puts vertices lying on many common meta-graphs close to each other in the embedding space.

\section{Meta-Graph Assessment}
\label{sec:assessment}
While meta-graphs are widely used for HIN modeling, different meta-graphs encode diverse semantics that essentially leads to rather different utilities, which might be understood by looking into the structures of the underlying projected networks \cite{li2016transductive,jiang2017semi}. To this end, we present our spectral embedding method, which naturally serves as a great tool to facilitate such assessment in an efficient way.

We notice that in spectral graph theory, eigenvalues are closely related to many essential graph properties \cite{chung1997spectral}. However, it is unknown what properties are indeed impactful, \ie, important for meta-graph utilities, especially regarding HIN embedding. To understand this, we conduct a systematic empirical study on various real-world HINs towards multiple traditional network mining tasks. Specifically, for each projected network, we visualize and study the correlations between its spectrum and embedding quality. As we will soon see, the results are indeed highly interpretable and insightful.

The datasets we use include HINs in different domains, \ie, DBLP from an academic publication collection\footnote{https://dblp.uni-trier.de/}, IMDB from a movie rating platform\footnote{http://www.imdb.com/}, and Yelp from a business review website\footnote{https://www.yelp.com/}. 
Details of these datasets are as follows.
%USPatent from a patent tracking service\footnote{https://www.uspto.gov/}, 

\begin{enumerate}
\item \textbf{DBLP}: We use the Arnetminer dataset V8\footnote{https://aminer.org/citation} collected by \cite{tang2008arnetminer}.
It contains four types of vertices, \ie, author (A), paper (P), venue (V), and year (Y). 
%The edge types include authors writing papers, papers citing papers, papers containing terms, papers publishing in venues, and papers publishing in years. 
%\item \textbf{USPatent}: We use the patent dataset of United States Patent and Trademark Office (USPTO)\footnote{http://www.patentsview.org/download/}.
%Following \cite{fu2017hin2vec}, we extract patents issued between 1998 to 2012 in 14 drug related classes to form a network, which contains patents (P), inventors (I) and assignees (A). Patent classes are used as labels for task evaluations. The edge types include inventorships (P-I), patents' assignees (P-A) and patent citations (P-P). 
\item \textbf{IMDB}: We use the MovieLens-100K dataset\footnote{https://grouplens.org/datasets/movielens/100k/} made public by \cite{harper2016movielens}. 
There are four types of vertices, \ie, user (U), movie (M), actor (A), and director (D). 
%The edge types include users reviewing movies, actors featuring in movies and directors making movies. 
\item \textbf{YELP}: 
We use the public dataset from the Yelp Challenge Round 11\footnote{https://www.yelp.com/dataset}. 
Following an existing work that models the YELP data with heterogeneous networks \cite{zhao2017meta}, we extract five types of vertices, \ie, business (B), user (U), location (L), category (C), and star (S). 
%The edge types include users reviewing businesses, businesses belonging to categories, businesses residing in locations, businesses having average stars, category related to categories and users being friends with users. 
\end{enumerate}

\subsection{FPP (First-Positive-Point) - Network Connectivity}
\begin{figure*}[h!]
\centering
\subfigure[Spectrum -- DBLP]{
\includegraphics[width=0.3\textwidth]{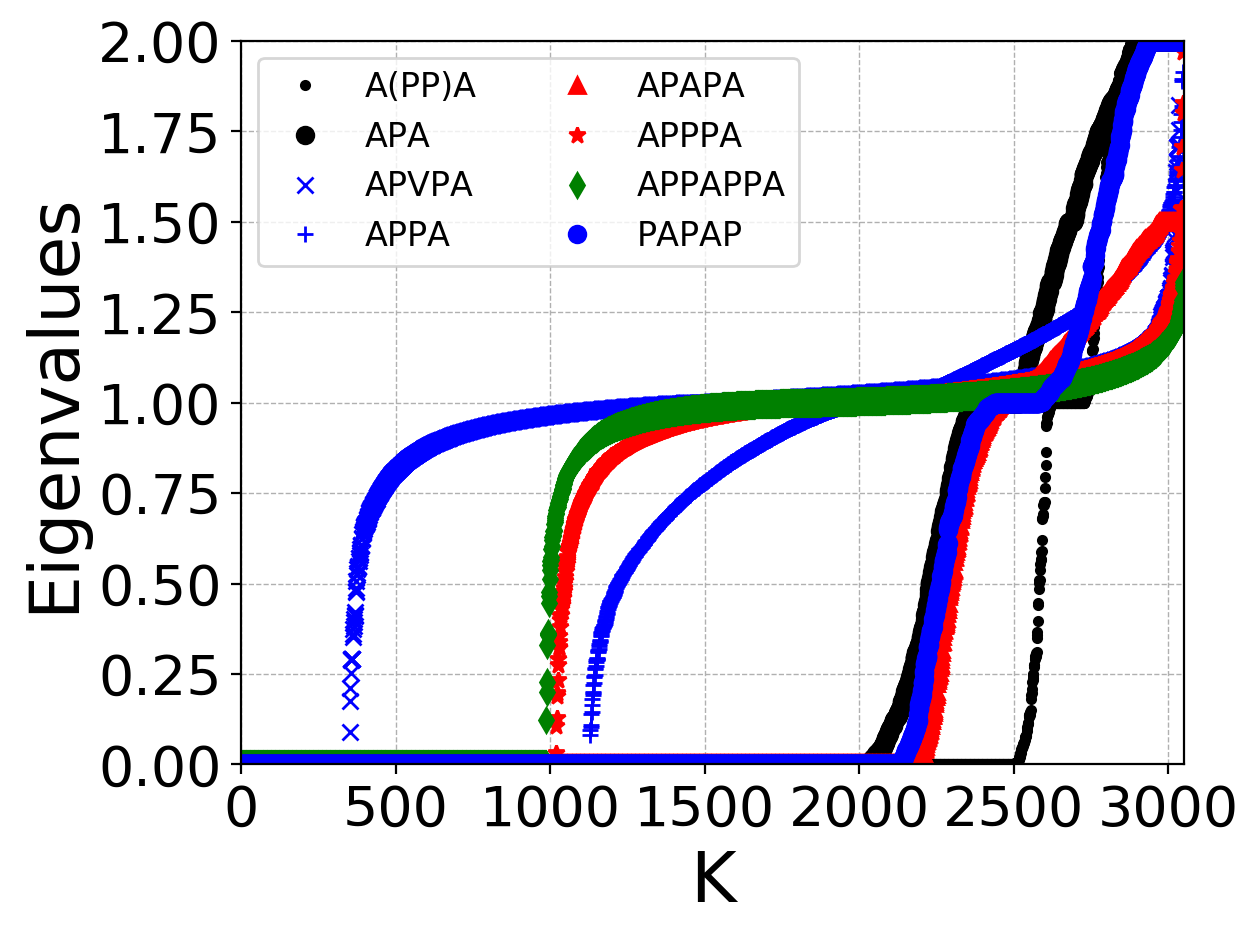}}
\subfigure[Spectrum -- IMDB]{
\includegraphics[width=0.3\textwidth]{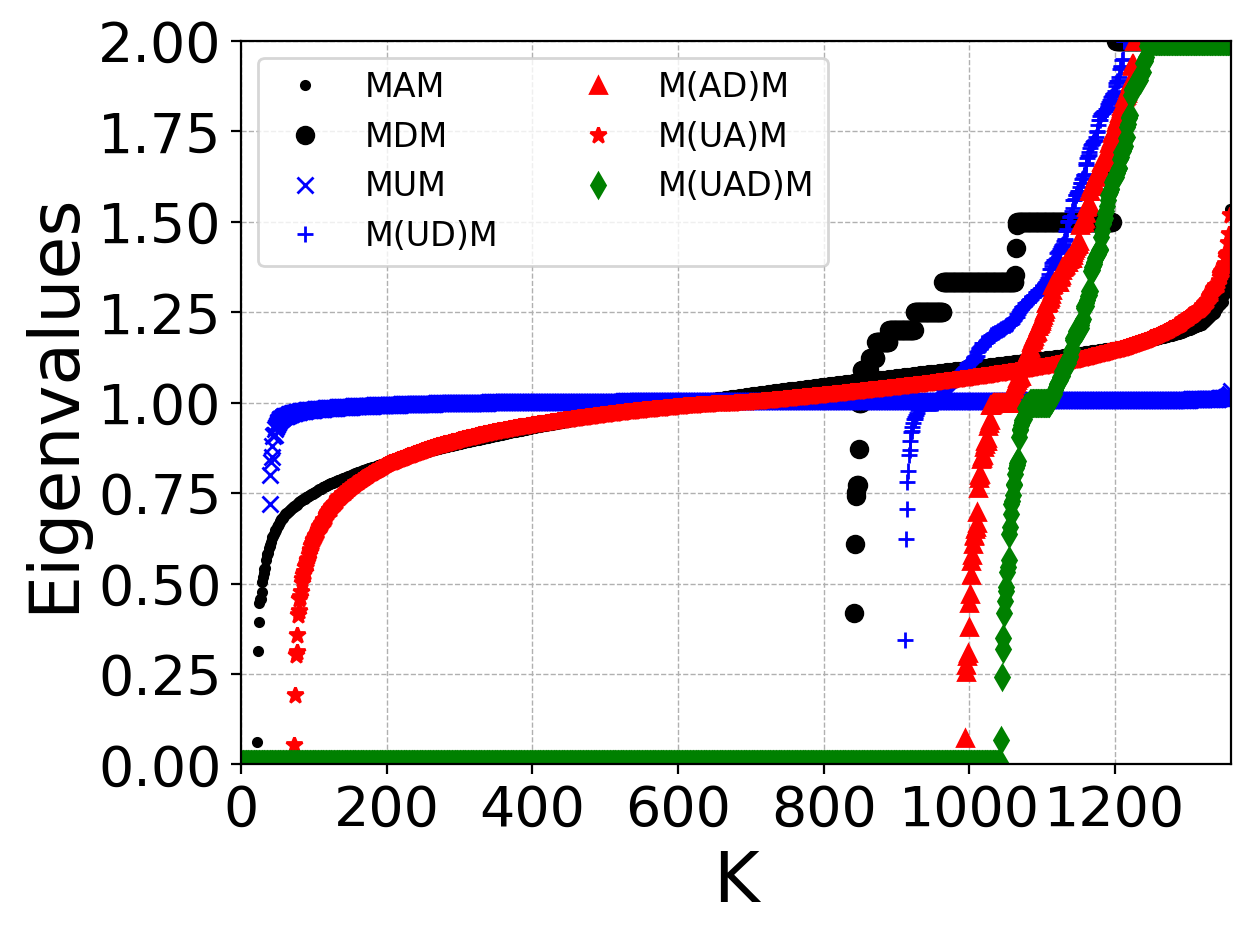}}
\subfigure[Spectrum -- Yelp]{
\includegraphics[width=0.3\textwidth]{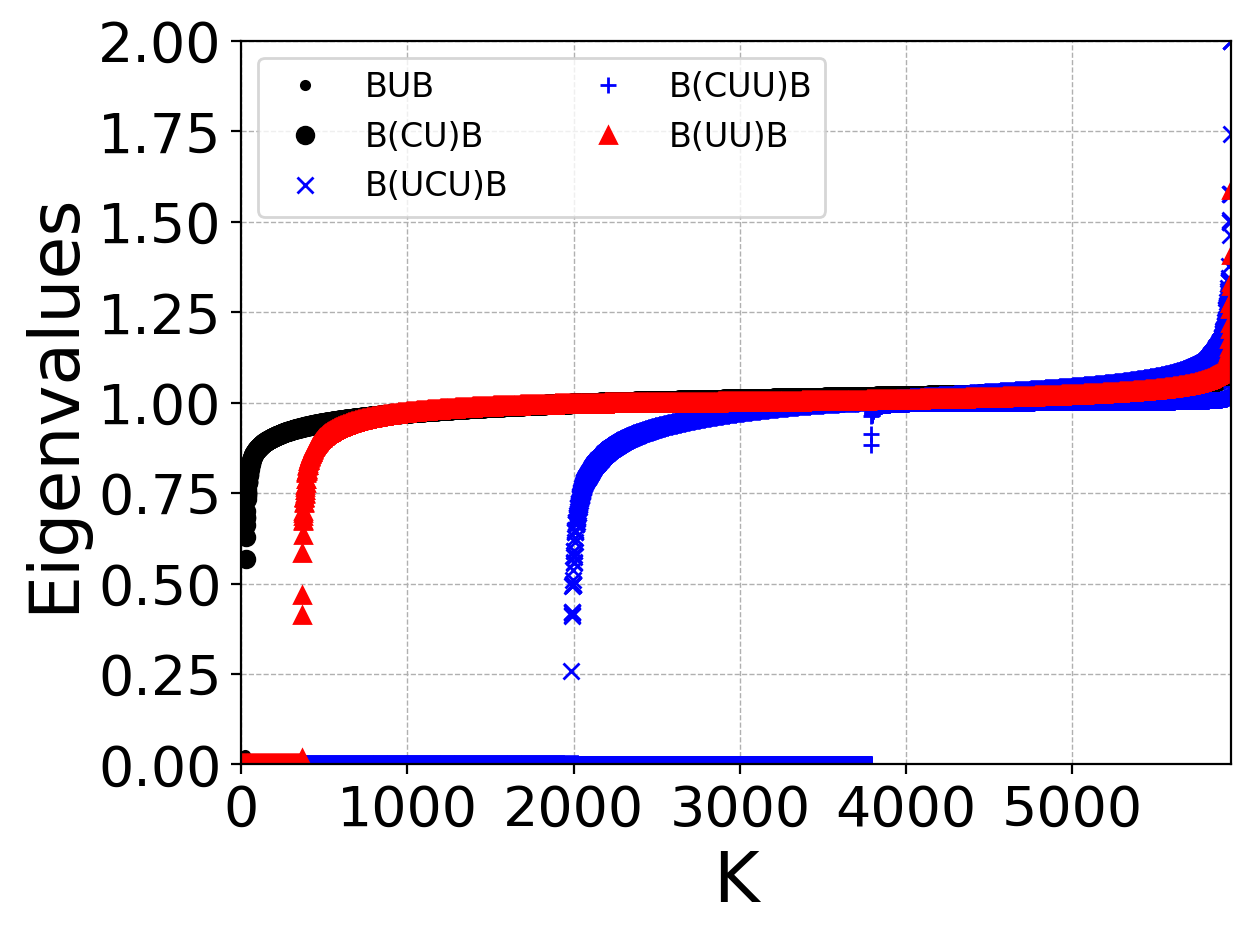}}
\subfigure[Performance -- DBLP]{
\includegraphics[width=0.3\textwidth]{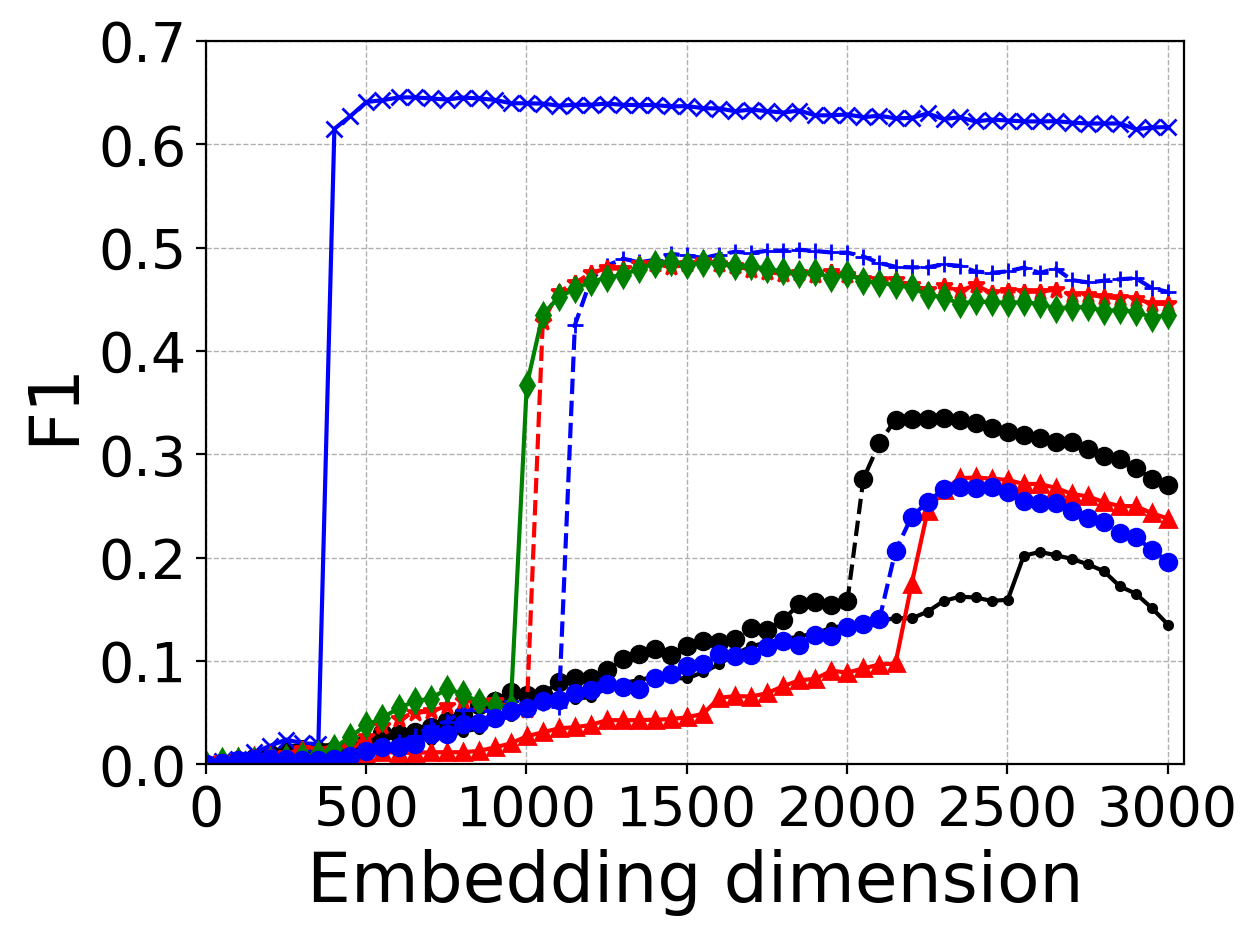}}
\subfigure[Performance -- IMDB]{
\includegraphics[width=0.3\textwidth]{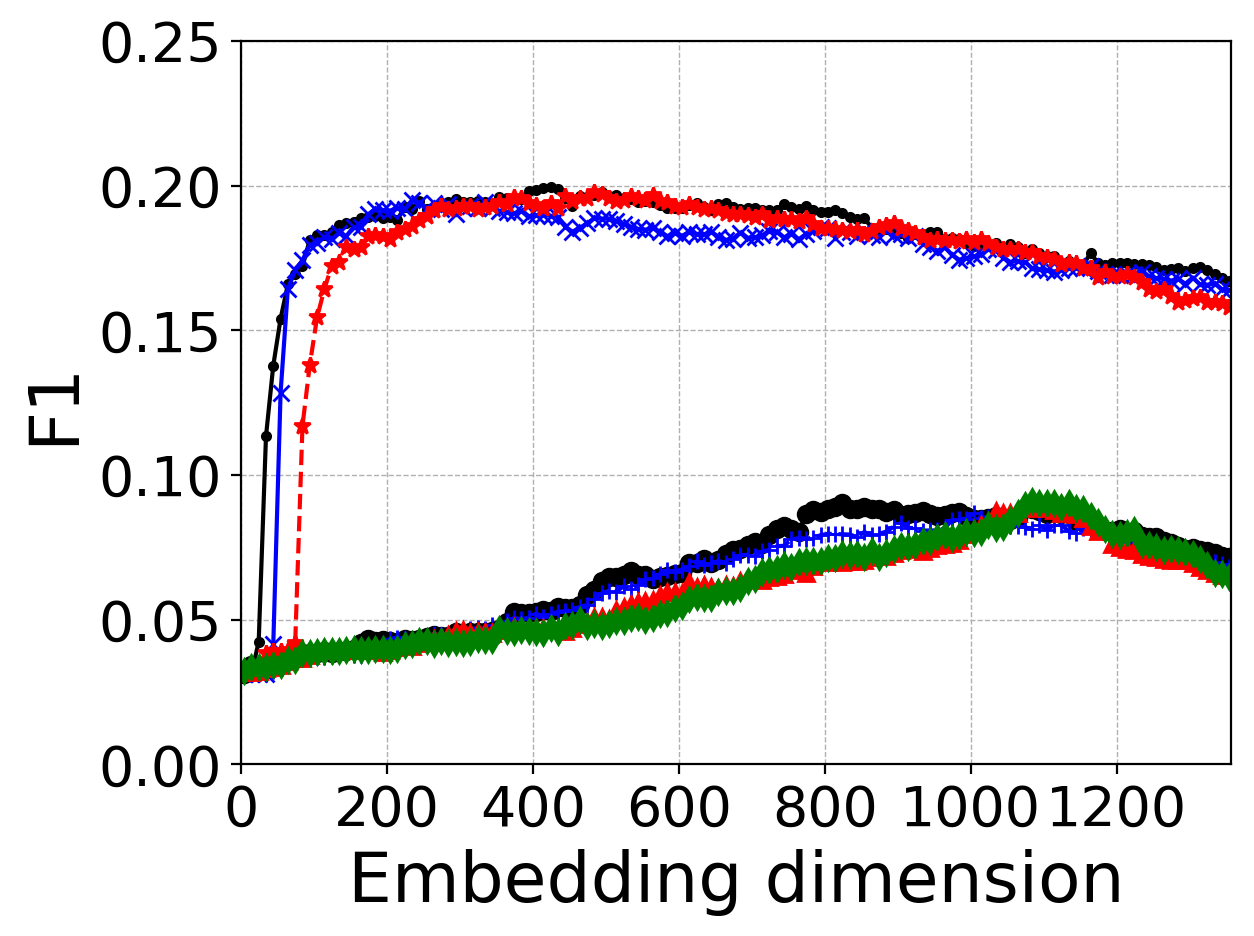}}
\subfigure[Performance -- Yelp]{
\includegraphics[width=0.3\textwidth]{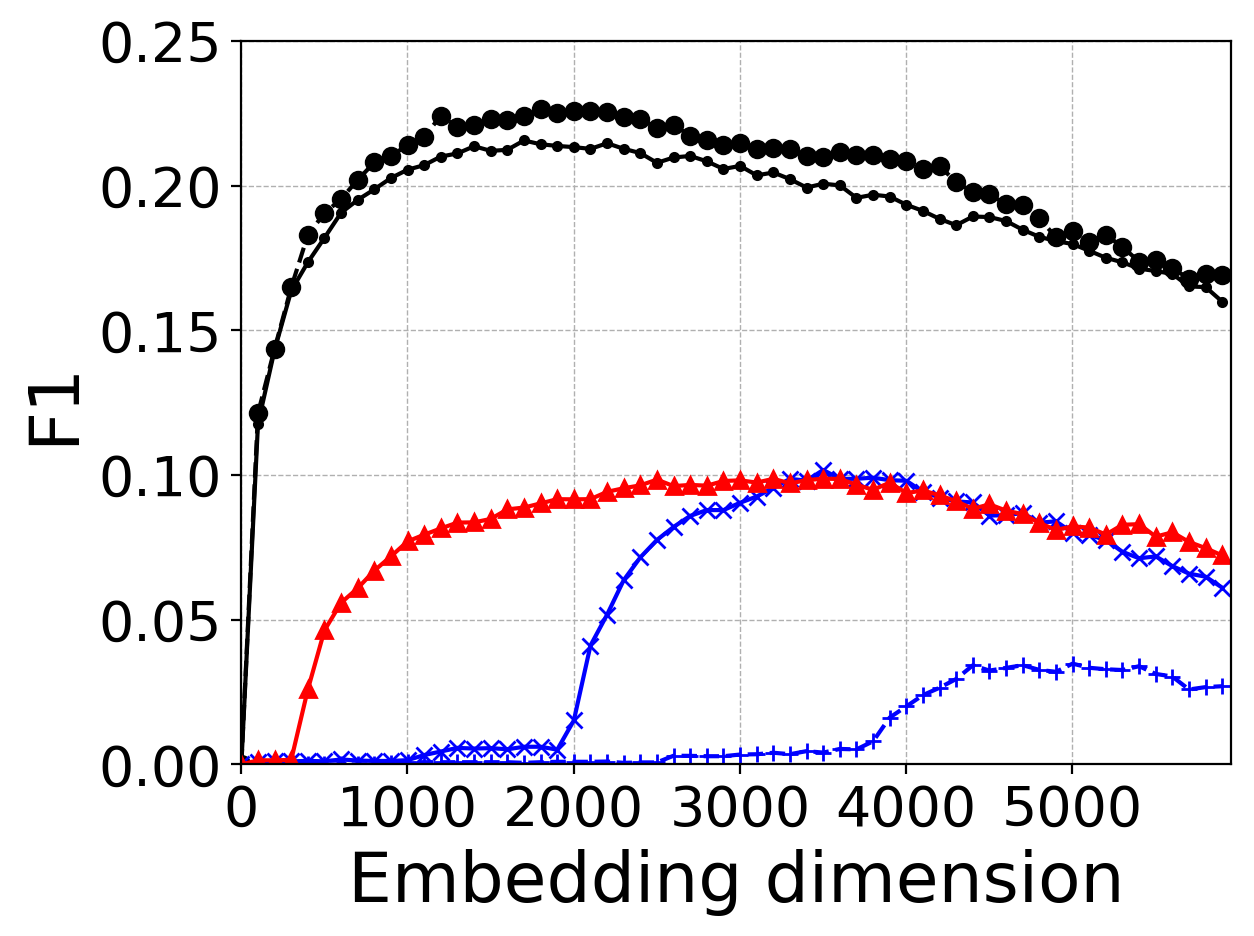}}
\caption{FPP of the spectra clearly correlates with the embedding performance. }
\label{fig:fpp}
\end{figure*}

\vspace{-5pt}
\header{Empirical Observations.}
Figure \ref{fig:fpp} shows the spectrum and embedding quality of different meta-graphs on the three datasets. The spectrum is computed via SVD\footnote{https://docs.scipy.org/doc/numpy.linalg.svd.html} on the normalized Laplacian defined in Eq.~\ref{eq:lap} and sorted in ascending orders. The embedding quality is evaluated towards node classification through an off-the-shelf SVM\footnote{http://scikit-learn.org/stable/modules/svm.html} model with standard five-fold cross validation on labeled nodes. We compute the commonly used $F1$ score for evaluating the classification performance. Other tasks like standard link prediction and clustering show similar trends and are omitted due to space limit.

As we can observe, the spectrum curve always starts from zero, and increase to positive values at some point, which we refer to as \textit{FPP (First-Positive-Point)}. Its position has a clear correlation with the embedding quality, \ie, (1) the spectrum curve and performance curve mostly start to grow at the same point, and (2) the earlier the spectrum curve starts to grow, the higher the performance curve can reach. 

\vspace{-5pt}
\header{Theoretical Interpretations.}
Looking at Figure \ref{fig:fpp} from a graph theory point of view, we find the strong correlations quite revealing.
According to Theorem \ref{thm:conn}, the number of zeros in the spectrum is exactly the number of disconnected components in the corresponding network. Hence, the results in Figure \ref{fig:fpp} clearly indicate that meta-graphs leading to better connected projected networks usually have better HIN embedding quality. 
The second observation is more interesting. Again, according to Theorem \ref{thm:conn}, the first several embedding dimensions that correspond to zero eigenvalues actually work as the features that identify whether the corresponding vertex belongs to a single connected component. As the number of embedding dimensions increases from 0 to FPP, the performance hardly improves, which implies that the identity of each connected component might be not useful for the HIN embedding. This observation is quite opposite to the recent significant findings in homogeneous networks \cite{benson2016higher,li2017inhomogeneous}, where in practice, with a good high-order connectivity pattern, the identity of connected component itself may have already been a strong feature for vertex embedding. 

The results further show that a small number of eigenvectors associated with small non-zero eigenvalues may help greatly towards the overall HIN embedding performance. According to Theorem~\ref{thm:nodal-domain}, within each connected component, these newly added eigenvectors begin to characterize the nodal domains within the connected components. Theorem~\ref{thm:high-order-cheeger} further implies that these nodal domains are essentially good network communities (\ie, densely connected parts) within the connected components. 
Therefore, the results can be understood as most of the connected components hold good community structures within themselves, and thus these components can be well represented by only a few eigenvectors associated with the small positive eigenvalues right after the FPP.

%\textcolor{blue}{The identity of community has little help while the low-harmonic structure of each connected component leads to good embedding performance. }
%\textcolor{red}{Carl: Pan, please help revise this paragraph.}
%Looking at Figure \ref{fig:all} from a graph theory point of view, we find the strong correlations not coincidental. Singularity of the spectrum curve exactly indicates the connectivity of the underlying network-- the number of zeros in the spectrum is the number of disconnected components in the corresponding network. It tells us that different meta-graphs do organize the HINs in different ways, but such organizations do not necessarily align with labeled data. It is necessary that we look into each natural clusters of the induced networks to find out the real semantically useful clusters. 
\vspace{-5pt}
\header{Efficient Assessments.}
Based on the systematic empirical study and theoretical interpretations, we are able to efficiently assess the meta-graph utility regarding HIN embedding by simply looking at the leading eigenvalues of the corresponding projected network. Particularly, meta-graphs corresponding to early FPP of the spectrum curves are generally more useful. Moreover, spectral embeddings corresponding to the positive eigenvalues are more important than those of the zero ones. 

\subsection{Curvature - Network Low-Rank Property}
\begin{figure}[h!]
\vspace{5pt}
    \centering
        \includegraphics[width=0.9\linewidth]{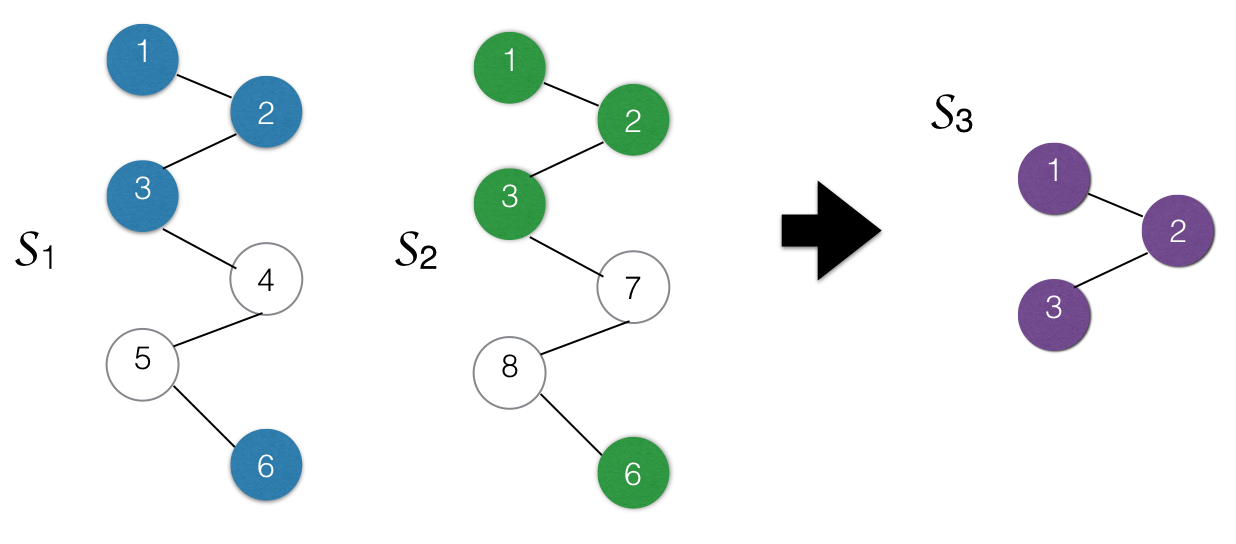}
        \vspace{-5pt}
    \caption{The spectra aligning process of finding LC3 of two projected networks.}
    \label{fig:common}
 \end{figure}

\vspace{-5pt}
\header{Empirical Observations.}
Besides FPP, is any other spectrum property indicative to the embedding quality?
To rule out the influence of FPP, we focus on each pair of meta-graphs and pick out their \textit{LC3 (Largest Common Connected Component)} as illustrated in Figure \ref{fig:common}. 
Suppose subnetworks $\mathcal{S}_1$ and $\mathcal{S}_2$ are the connected components on the projected networks of meta-graph $\mathcal{M}_1$ and $\mathcal{M}_2$, respectively. Then subnetwork $\mathcal{S}_3$ is a common connected component of $\mathcal{M}_1$ and $\mathcal{M}_2$. LC3 is the one with the largest number of vertices. On LC3, the spectra of two meta-graphs are aligned, in a way that they both only have a single zero value.

\begin{figure*}[t!]
\centering
\subfigure{
\includegraphics[width=0.3\textwidth]{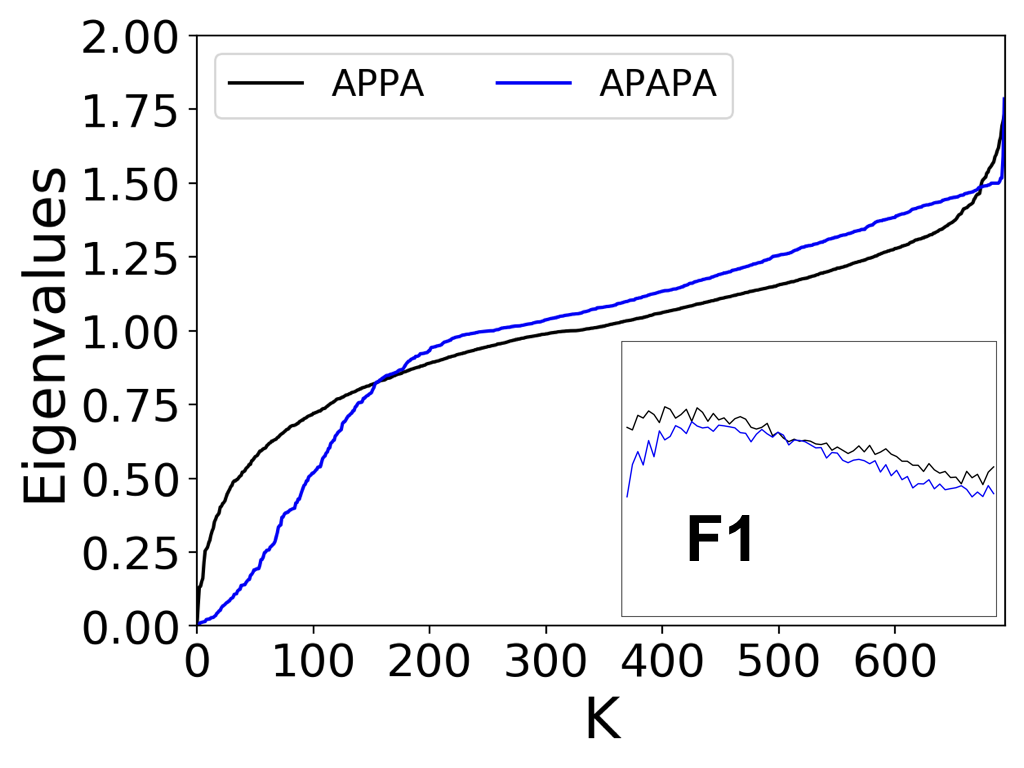}}
\subfigure{
\includegraphics[width=0.3\textwidth]{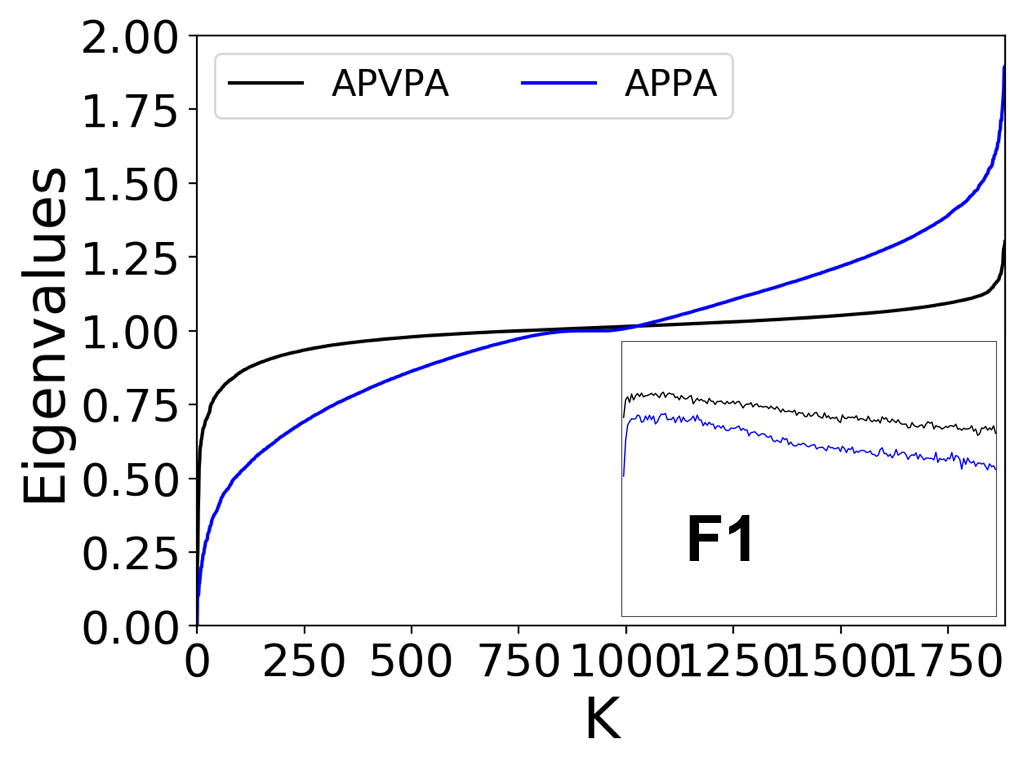}}
\subfigure{
\includegraphics[width=0.3\textwidth]{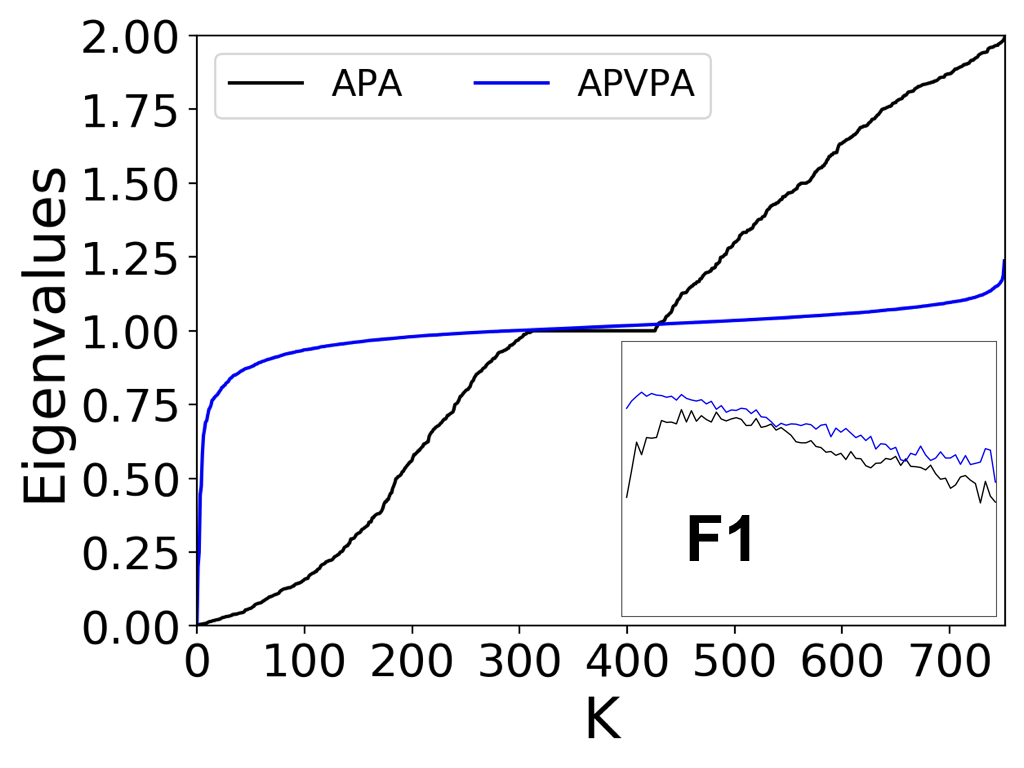}}

%\subfigure[Case 1: Performance]{
%\includegraphics[width=0.3\textwidth]{figures/pair_performance_1.png}}
%\subfigure[Case 2: Performance]{
%\includegraphics[width=0.3\textwidth]{figures/pair_performance_2.png}}
%\subfigure[Case 3: Performance]{
%\includegraphics[width=0.3\textwidth]{figures/pair_performance_3.png}}
\vspace{-10pt}
\caption{The curvature of the spectrum clearly correlates with the embedding performance.}
\label{fig:pair}
\end{figure*}

Throughout our systematic empirical study, we find the spectrum curvature highly correlated with the embedding quality. Particularly, as shown in Figure \ref{fig:pair}, (1) the faster the eigenvalues grow in the beginning (\ie, larger curvature), the better the embedding quality is, and (2) the embedding quality degenerates with larger embedding sizes. Although we only present the performance towards author classification on DBLP due to space limit, we find exactly similar phenomena for other network mining tasks on all three datasets we use.

\vspace{-5pt}
\header{Theoretical Interpretations.} 
The observations again can be interpreted through references to spectral graph theory. First, better embedding based on the faster growth of eigenvalues can be explained from the perspective of PCA. With simple linear algebra, we know the optimal loss of PCA (Eq.~\ref{eq:pca}) equals to $\sum_{i=k+1}^{|\mathcal{V}|} (2-\lambda_i)^2$. The fast growth of eigenvalues means for some small $i$, $\lambda_{i}$ can be already large and hence $2-\lambda_i$ is small. One step further, it implies that the projected network has preferable low-rank properties, 
\ie, a steep curvature indicates the energy mostly concentrates on a few eigenvalues of the normalized adjacency matrix $\mathcal{A}$. 
Therefore, the loss of PCA can be small for some small embedding dimension $k$, and $\mathcal{A}$ can be well approximated by the inner product of low-dimensional embedding vectors of different vertices, \ie, $HWH^T$. When the eigenvalues achieve the medium value (almost 1), the nodal domains of the corresponding eigenvectors can hardly express the community structures of the network according to the inequality in Eq.~\ref{eq:high-order-cheeger}, as the RHS is greater than 1 and becomes trivial. Therefore, the eigenvectors \wrt~large eigenvalues ($>$1) may not be a good representation that encodes the topology of the network. As a consequence, eigenvectors of large $\lambda_i$ are not informative for the HIN embedding. In fact, adding them as the embedding features may cause significant overfitting and hence the degenerated learning performance.

\vspace{-5pt}
\header{Efficient Assessments.}
Besides FPP, the curvature of spectrum allows additional efficient assessment of the meta-graph utility, \ie, steeper eigenvalue growth indicates better embedding performance. Moreover, spectral embeddings corresponding to the first several non-zero eigenvalues carry the most useful structure information, while the subsequent ones are less useful and may easily lead to model overfitting. 

\subsection{Implied Assessment Method}
Given a meta-graph $\mathcal{M}$ on an arbitrary HIN $\mathcal{N}$, without knowing the downstream task, we simply need to compute the projected network and its leading eigenvalues, based on which we can then quickly assess the utility of $\mathcal{M}$ and select the most informative embedding dimensions. Note that, this method is efficient due to several facts: (1) Given $\mathcal{N}$ and $\mathcal{M}$, it is not always necessary to compute the projected networks from scratch. In fact, many real-world network companies nowadays maintain the graph databases to constantly track and store the instances of certain high-order structures for analytical usage \cite{angles2008survey}. (2) Finding instances of $\mathcal{M}$ on $\mathcal{N}$ is a well-studied problem, which can be efficiently solved by algorithms like \cite{fang2016semantic}. (3) Our method only works with the projected networks, which are much smaller than the original HINs. (4) We only need to check the leading $K$ eigenvalues and do not require the networks to be fully decomposed \cite{davidson1975iterative}.

%!TEX root = hinse.tex
\section{Meta-Graph Combination}

\label{sec:combination}
\subsection{Motivations and Challenges}
\begin{figure}[h!]
    \centering
        \includegraphics[width=0.9\linewidth]{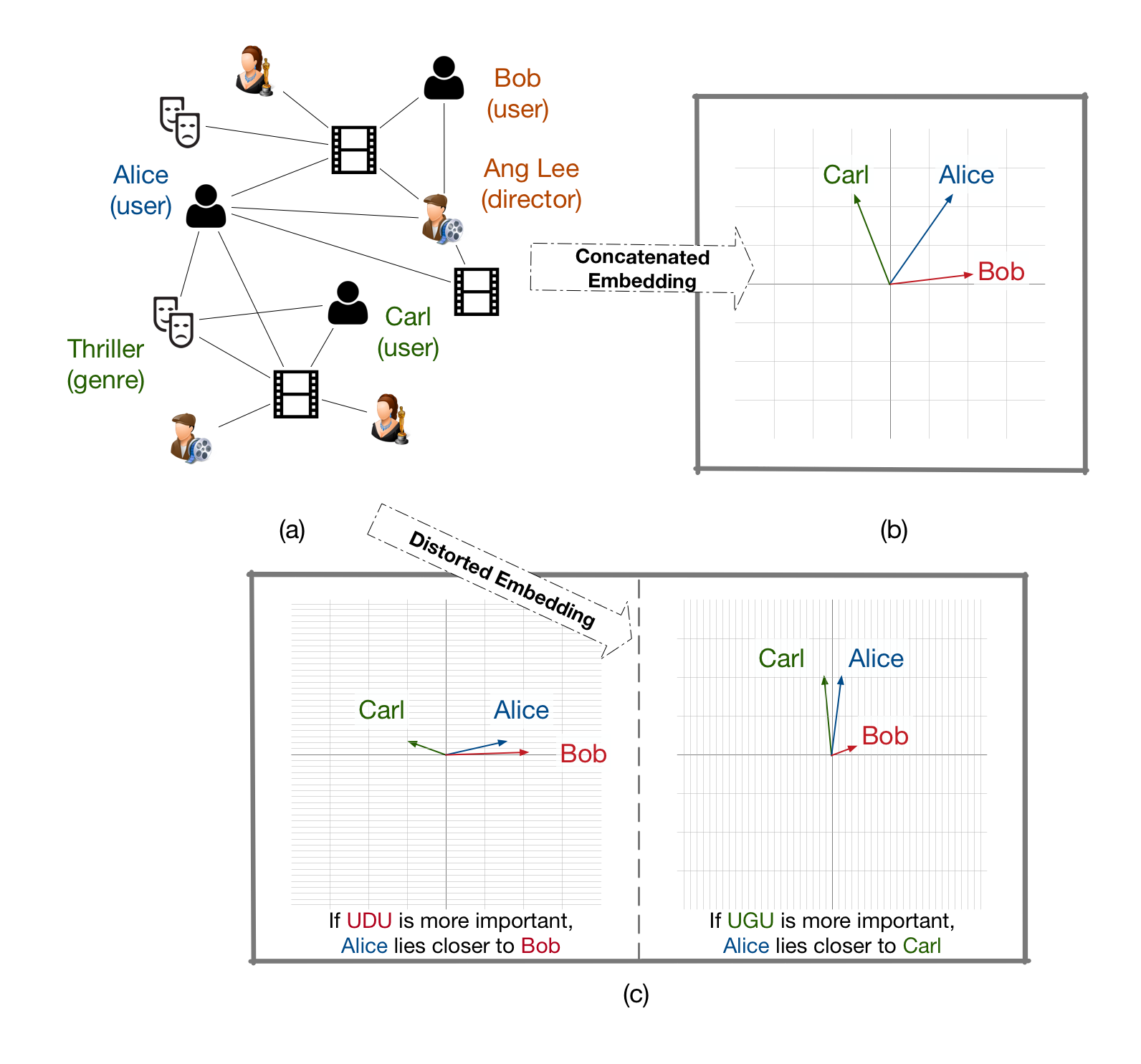}
        \vspace{-20pt}
    \caption{A toy example of a movie-review HIN.}
     \vspace{-5pt}
    \label{fig:movie}
 \end{figure}

In HIN, each meta-graph captures particular semantics. Take Figure \ref{fig:movie} (a) as an example. In a movie-review HIN, suppose \textit{Alice} is connected with \textit{Bob} by meta-graph $\mathcal{M}_1$ (\textit{UDU}), and with \textit{Carl} by $\mathcal{M}_2$ (\textit{UGU}). Thus, the underlying semantics are, \textit{Alice} and \textit{Bob} like movies directed by the same \textit{director}, while she and \textit{Carl} like movies of the same \textit{genre}.
For the general purpose of HIN embedding, it is natural that we want the embeddings to capture all ``useful'' semantics by simultaneously considering multiple meta-graphs.

However, simply concatenating the individual embeddings of multiple meta-graphs may actually lead to poor results. To illustrate this, we continue with the example in Figure \ref{fig:movie}. In the concatenated embedding space of $\mathcal{M}_1$ and $\mathcal{M}_2$, \textit{Bob} and \textit{Carl} might be far away, since they do not like the same movies. As a consequence, \textit{Alice} can only lie between the two of them while being close to neither of them, due to the triangle inequality property of metric spaces as shown in (b). It implies that, in order to capture the essentially useful semantics, we need to wisely distort the embedding space, by throwing away redundant, noisy and non-discriminative information. Eventually, we want a model that is able to automatically trade-off different meta-graphs and their embedding dimensions, and arrive at an embedding space like one of those in (c), where \textit{Alice} can be close to either \textit{Bob} or \textit{Carl}, depending on which meta-graph is found to be more important.
 
We find that the problem of unsupervised meta-graph combination essentially boils down to two challenging subproblems as follows:
\begin{enumerate}
\item Dimension Reduction: 
As we have just explained, simply concatenating the individual embeddings ignores the interactions and correlations among meta-graphs, and results in high dimensionality and data redundancy. Moreover, as we can observe from our analyses in Section \ref{sec:assessment}, the individual embeddings can be quite noisy, which together with the high dimensionality can easily lead to model overfitting.
\item Meta-graph Selection: As we also observe in Section \ref{sec:assessment}, the utilities of meta-graphs towards HIN embedding can be rather different. While they can be efficiently assessed individually, there is no end-to-end systematic method for the selection of important meta-graphs by considering them together in an unsupervised way, so as to capture all essentially useful semantics in a HIN.
%\item Efficient Training: While many methods exist for dimension reduction based on matrix decomposition, they are more efficient for the linear models, and become less applicable with real-world large networks with millions of vertices. 
\end{enumerate}

\subsection{Autoencoder with $\ell_{2,1}$-Loss}
\begin{figure}[h!]
    \centering
        \includegraphics[width=0.9\linewidth]{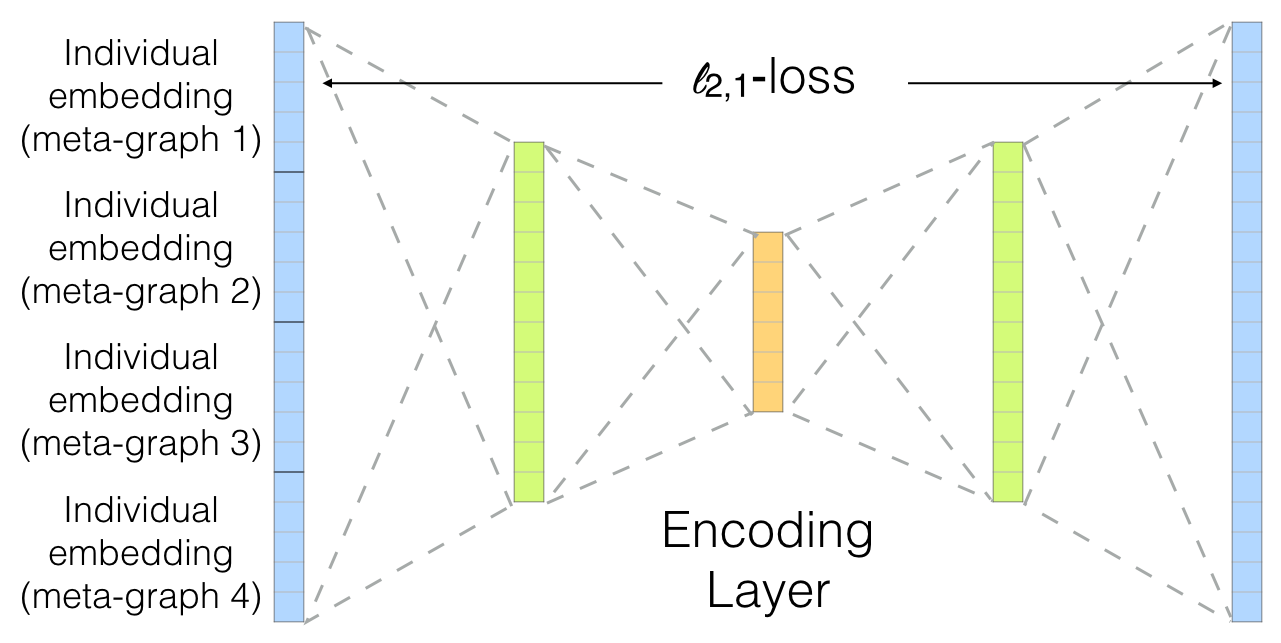}
        \vspace{-10pt}
    \caption{Our joint embedding framework for meta-graph combination.}
    \label{fig:com}
 \end{figure}

To simultaneously solve the above two problems, we propose the method of autoencoder with $\ell_{2,1}$-loss. The overall framework is shown in Figure \ref{fig:com}.

For unsupervised dimension reduction, we take the spirit of \cite{he2006laplacian,wold1987principal} in preserving the most representative features by variance maximization. Further, we get motivated by recent advances in neural networks and deep learning, particularly, the unsupervised deep denoise autoencoders \cite{vincent2008extracting,le2013building}. They have been shown effective in feature composition due to the proven advantages in capturing the intrinsic features within high-dimensional noisy redundant inputs in a non-linear way. 

One step further, we design a specific $\ell_{2,1}$-loss to further require grouped sparsity on the embedding dimensions \wrt~each meta-graph, so as to effectively select the more useful meta-graphs in an end-to-end fashion. It helps us to put more stress on the important meta-graphs to improve the final embedding quality. Moreover, it also enables better understanding of the meta-graph utilities, and allows further validation of our meta-graph assessment methods. 

In what follows, we go through our model design in details.

%As we will show later, the encoded features produced by autoencoders can be regarded as a non-linear combination of the most representative original features in terms of variance, and such representativeness can be theoretically guaranteed in comparison with the original features.

For each vertex $v_i \in \mathcal{V}$, given its spectral embedding of the $k$-th projected network $\mathbf{h}^k_i$, the input of our meta-graph combination framework is thus $\mathbf{x}_i=\Phi(\mathbf{h}^1_i, \ldots, \mathbf{h}^K_i)$, which is a vector concatenation of $K$ spectral embeddings. 

To leverage the power of autoencoders, given $\mathbf{x}_i$, we first apply an encoder, which consists of multiple laters of fully connected feedforward neural networks with \textit{LeakyReLU} activations. The neural networks are in decreasing sizes and after them we get a $Q$-dim compressed embedding $\mathbf{q}_i$ as
\begin{align}
\mathbf{q}_i=\mathbf{f}_e^P(\ldots \mathbf{f}_e^2(\mathbf{f}_e^1(\mathbf{x}_i))\ldots),
\end{align}
where $P$ is the number of hidden layers in the encoder, and
\begin{align}
\mathbf{f}_e^p(\mathbf{x}) = \text{LeakyReLU}(\mathbf{W}_e^p \text{Dropout}(\mathbf{x})+\mathbf{b}^p_e).
\end{align}
 %$\Theta_e$ is the set of parameters in the $H$ encoder layers.

To ensure that $\mathbf{q}_i$ captures the important information in $\mathbf{x}_i$, we compute the reconstruction $\mathbf{\tilde{x}}_i$ of $\mathbf{x}_i$ through stacking a decoder, which also consists of multiple layers of fully connected feedforward neural networks. The sizes of neural networks are in an increasing order, exactly the opposite as in the encoder.  So we have
\begin{align}
\mathbf{\tilde{x}}_i=\mathbf{f}_d^P(\ldots \mathbf{f}_d^2(\mathbf{f}_d^1(\mathbf{q}_i))\ldots),
\end{align}
\begin{align}
\mathbf{f}_d^p(\mathbf{x}) = \text{LeakyReLU}(\mathbf{W}_d^p \text{Dropout(}\mathbf{x})+\mathbf{b}^p_d).
\end{align}

The number of hidden layers in the decoder is also $H$, the same as in the encoder. %$\Theta_d$ is the set of parameters in the $H$ decoder layers.

After the decoder, a reconstruction loss for embedding is computed as
\begin{align}
\mathcal{J}=\sum_{i=1}^n l(\mathbf{x}_i,\mathbf{\tilde{x}}_i),
\label{eq:loss}
\end{align}
which is a summation over all vertices in $\mathcal{V}$. 

For regular autoencoders, $l$ is implemented either as a cross entropy for binary features, or a mean squared error for continuous features. However, per reasons we have just discussed in Section \ref{sec:combination}.A, we apply a specific $\ell_{2,1}$-loss \cite{ma2017multi} to $l$ as
\begin{align}
l(\mathbf{x}_i,\mathbf{\tilde{x}}_i)&=||\mathbf{x}_i-\mathbf{\tilde{x}}_i||_{2,1}\\
&=\sum_{k=1}^K ||\mathbf{h}_i^k-\mathbf{\tilde{h}}_i^k||_2,
\end{align}
where $\mathbf{\tilde{h}}_i^k$ is the reconstructed embedding of $\mathbf{h}_i^k$.

\subsection{Theoretical Justification} 
Autoencoder is a non-linear generalization of PCA. Particularly, consider an $\ell_2$-loss in Eq.~\ref{eq:loss}. It is exactly the same as the PCA loss in Eq.~\ref{eq:pca}, if we remove the non-linear activation layers. From the mathematical geometry point of view, consider the original embedding space as a ball, PCA distorts this ball into an ellipsoid by picking out the directions of the greatest variance in the dataset. This process necessarily incurs an information loss, but the variance maximization process ensures the lost information to be more of the redundant part. 

One step further, our leverage of autoencoder further enables the utilization of the expressiveness of non-linear feedforward neural networks. It allows us to efficiently explore more complex interactions of different embedding dimensions and distort the embedding space with more flexibility \cite{hinton2006reducing}.

Beyond the standard autoencoder, our $\ell_{2,1}$-loss is built on group-wise feature selection via group lasso~\cite{friedman2008sparse}. The setting of the $\ell_1$-loss only imposes sparsity in the group level while the $\ell_2$-loss within each group expresses that all features of one group should be selected or rejected simultaneously. 
The mathematical property of $\ell_{2,1}$-loss coincides with our target of combining the individual embeddings of different meta-graphs: When compressing the embeddings, we want the model to ensure that only some grouped dimensions are exactly reconstructed, which allows it to ignore certain useless meta-graphs and instead focus on the more important ones.  
%Also, there should be less difference among the embedding dimensions induced by the same meta-graph. 
In this way, our model is able to select important meta-graphs in an end-to-end fashion.

%Finally, the model can be trained via standard automatic stochastic gradient descent, and further optimized via well-developed techniques like mini-batch Adam \cite{kinga2015method} and layer-by-layer pre-training \cite{hinton2006reducing}.

\subsection{Empirical Analysis} 
We conduct a series of empirical analyses to specifically study our meta-graph combination model. The autoencoder we use in this subsection has only one encoding layer with no additional hidden layer. For input, we take an 80-dimensional individual spectral embedding for each meta-graph. Due to space limit, we focus the analyses on the DBLP dataset with four meta-graphs: \textit{APVPA, APPA, APPPA} and \textit{APPAPPA}.
% Current AE structure is 40*4 (160) -> 160 -> Encoding
%\begin{table}[h!]
%\centering
% \begin{tabular}{|c||c|c|c|c|c|}
%   \hline
%{\bf $|Q|$}&25&35&50&65&85\\
%  \hline
%  \hline
%{\bf Linear}& 0.647 & 0.631 & 0.586 & 0.603 & 0.618 \\
%  \hline
%{\bf Non-linear}& 0.669 & 0.664 & 0.675 & 0.659 & 0.668 \\
%\hline
% \end{tabular}
%  \vspace{2pt}
% \caption{ \label{tab:linear}Comparing the F1 scores of linear and non-linear models.}
% \vspace{-10pt}
%\end{table}

\begin{table}[h!]
\centering
 \begin{tabular}{|c||c|c|c|c|c|c|}
   \hline
{\bf $|Q|$}&10&20&40&70&100&200\\
  \hline
  \hline
{\bf Linear}& 0.582 & 0.611 & 0.628 & 0.631 & 0.637 & 0.642\\
  \hline
{\bf Non-linear}& 0.654 & 0.668 & 0.673 & 0.669 & 0.668 & 0.676\\
\hline
 \end{tabular}
  \vspace{2pt}
 \caption{ \label{tab:linear}Comparing the F1 scores of linear and non-linear models.}
 \vspace{-10pt}
\end{table}

\begin{table}[h!]
\centering
 \begin{tabular}{|c||c|c|c|c|c|}
   \hline
{\bf Meta-graphs}&\textit{APVPA}&\textit{APPA}&\textit{APPPA}&\textit{APPAPPA}&F1\\
  \hline
  \hline
{\bf $\ell_{2,1}$}& 0.214 & 0.256 & 5.222 & 5.243 & 0.695 \\
  \hline
{\bf $\ell_{2}$}& 2.420 & 3.133 & 3.349 & 3.340 & 0.668 \\
\hline
 \end{tabular}
  \vspace{2pt}
 \caption{ \label{tab:l21}Comparing autoencoders with $\ell_{2,1}$-loss and $\ell_{2}$-loss.}
 \vspace{-10pt}
\end{table}

The first analysis we did is on the comparison between linear and non-linear autoencoders. In Table \ref{tab:linear}, the \textit{non-linear} results are constantly better than the \textit{linear} ones, which is generated by the exact same architectures with the non-linear activation functions removed. The differences are more significant for smaller encoding dimensions. It clearly indicates the power of non-linear embedding and supports our selection of autoencoder as the basic model.

Subsequently, we study the efficacy of our $\ell_{2,1}$-loss. In Table \ref{tab:l21}, we compare two models with the $\ell_{2,1}$-loss and standard $\ell_{2}$-loss. The encoding dimension is fixed to 200, and the losses are all group-wise computed after vector mean-shifting and normalization. As we can see, the $\ell_{2,1}$-losses can effectively differentiate the utilities of \textit{APVPA} and \textit{APPA} from \textit{APPPA} and \textit{APPAPPA}, while the $\ell_{2}$-losses are more uniform over all meta-graphs. The final embedding quality regarding the classification F1 score with the $\ell_{2,1}$-loss is also significantly better. It confirms our intuition of leveraging the group lasso for end-to-end meta-graph selection. 

Moreover, such results from our combination method clearly deem the meta-graphs \textit{APVPA} and \textit{APPA} to be more important than \textit{APPPA} and \textit{APPAPPA}, which aligns with our assessment method in Section \ref{sec:assessment}. Such observation allows us to close the gap between these two methods, and further propose a unified framework for meta-graph based HIN embedding. To be specific, given a large number of candidate meta-graphs (due to the lack of precise domain knowledge), our assessment method can be firstly applied for an efficient but coarse selection of individual candidate meta-graphs as well as promising embedding dimensions. Then our combination method can be applied to fine-tune the combined embedding, which results in low-dimensional high-quality representations capturing the most important information across multiple meta-graphs. 

\section {Experimental Evaluation} 
\label{sec:exp}
We comprehensively evaluate the performance of our proposed method in comparison with various state-of-the-art NN-based HIN embedding algorithms on the same three large real-world datasets as we described in Section \ref{sec:assessment}.
Extensive experimental results show that our method can effectively select and combine useful meta-graphs for general-purpose unsupervised HIN embedding, which leads to supreme performance on multiple traditional network mining tasks.

\subsection{Experimental Settings}
\vspace{-2pt}
\header{Datasets.} The datasets we use are DBLP, IMDB and Yelp, as described in Section \ref{sec:assessment}, with statistics shown in Table \ref{tab:stat}.

\begin{table}[h!]
 \vspace{2pt}
\centering
 \begin{tabular}{|c||c|c|c|c|c|}
   \hline
{\bf Dataset}&{\bf Size}&{\bf \#Types}&{\bf \#Nodes}&{\bf \#Links}&{\bf \#Classes}\\
  \hline
  \hline
{\bf DBLP}& 4.33GB & 4 & 335,185 & 2,704,655 & 4 \\
  \hline
{\bf IMDB}& 16.1MB & 4 & 45,913 & 153,645 & 23 \\
\hline
{\bf Yelp}& 6.52GB & 5 & 1,123,649 & 8,912,736 & 6 \\
\hline
 \end{tabular}
 \vspace{2pt}
 \caption{ \label{tab:stat}Statistics of the four public datasets we use.}
 \vspace{-10pt}
\end{table}

\header{Baselines.}
We compare with various unsupervised HIN embedding algorithms to comprehensively evaluate the performance of our proposed method.
\begin{itemize}
\item \textbf{PTE} \cite{tang2015pte}: It decomposes the heterogeneous network into a set of bipartite networks and then captures first and second order proximities for HIN embedding.
\item \textbf{Meta2vec} \cite{dong2017metapath2vec}: It leverages heterogeneous random walks and negative sampling for HIN embedding.
\item \textbf{ESim} \cite{shang2016meta}: It leverages meta-path guided path sampling and noise-contrastive estimation for HIN embedding.
\item \textbf{HINE} \cite{huang2017heterogeneous}: It captures $w$-hop neighborhoods under meta-path constrained path counts fo HIN embedding.
\item \textbf{Hin2vec} \cite{fu2017hin2vec}: It jointly learns the node embeddings and meta-path embeddings through relation triple prediction.
\item \textbf{AspEm} \cite{shi2018aspem}: It selects meta-graphs based on a heuristic incompatibility score and combine the embedding of multiple induced graphs through vector concatenation.
\end{itemize}

\vspace{-8pt}
\header{Evaluation protocols.}
We study the embedding quality of all algorithms on two traditional network mining tasks, \ie, node classification and link prediction. The class labels and evaluation links are generated as follows.
For DBLP, we use the manual class labels of authors from four research areas, \ie, \textsf{database, data mining, machine learning} and \textsf{information retrieval} provided by \cite{sun2011pathsim}.
For IMDB, we follow \cite{shi2018aspem} to use all 23 available genres such as \textsf{drama, comedy, romance, thriller, crime} and \textsf{action} as class labels.
%For USPatent, we follow \cite{fu2017hin2vec} to use the14 popular patent class labels as including \textsf{surgery, optics, dentistry, prosthesis} and so on.
For Yelp, we extract six sets of businesses based on some available attributes, \ie, \textsf{good for kids, take out, outdoor seating, good for groups, delivery} and \textsf{reservation}. 
Following the common practice in \cite{fang2016semantic,meng2015discovering}, for each dataset, we assume that authors (movies, businesses) within each semantic class are similar in certain ways, and generate pairwise links among them for the evaluation of link prediction.

All algorithms learn the embeddings on the whole network. 
For node classification, we split the class labels in half for training and testing. We train a standard SVM\footnote{http://scikit-learn.org/stable/modules/svm.html} on the training data and compute the \textit{F1} and \textit{Jaccard} scores towards all class labels on the testing data.
%For network clustering, we apply standard k-means\footnote{http://scikit-learn.org/stable/modules/generated/sklearn.cluster.KMeans.html} and compute \textit{F1} and \textit{Jaccard} scores towards all class labels. 
For link prediction, we compute the cosine similarity of each node pair, and rank all nodes for each node to compute the \textit{precision at $K$} and \textit{recall at $K$}. 
All algorithms are run on a server with one GeForce GTX TITAN X GPU and two Intel Xeon E5-2650V3 10-core 2.3GHz CPUs.
While scalability is not our focus in this work, we also measure the training time of all algorithms. Our audoencoder-based embedding model can be efficiently trained on GPU and consumes no significantly more time than most baselines.

\header{Parameter settings.}
Our method only has a few parameters. The sizes of spectral embeddings are set \wrt~our assessment method (\ie, 80 for DBLP, 150 for IMDB and 800 for Yelp). For the autoencoder, we empirically set the number of both encoder and decoder layers to 2, each halving (doubling) the size of the previous layer. The drop out ratio is 0.2.

For each HIN, we firstly enumerate all meta-graphs up to size 5 and visualize their spectra to select a few most promising meta-graphs by our assessment method\footnote{IMDB: \textit{MDM}, \textit{MAM}, \textit{MUM}, \textit{M(UD)M}, \textit{M(AD)M}, \textit{M(UA)M}, \textit{M(UAD)M};  DBLP: \textit{A(PP)A}, \textit{APA},  \textit{APPA}, \textit{APVPA},  \textit{APAPA},  \textit{APPPA}, \textit{PAPAP},  \textit{APPAPPA}; Yelp: \textit{BUB}, \textit{B(UC)B}, \textit{B(UCU)B}, \textit{B(CUU)B}, \textit{B(UU)B}}. These meta-graphs are then given as input to our combination method. Since most promising meta-graphs are actually meta-paths, they are also given as input to all compared baselines. All other parameters of the baselines are either set as given in the original work on the same datasets, or tuned to the best through standard five-fold cross validation on each dataset.

\subsection{Performance Comparison with Baselines}
As we can see from Table \ref{tab:class} and \ref{tab:link}, the HIN embeddings produced by our method constantly lead to better performance on both node classification and link prediction tasks. The results on node classification are all averaged over 10 random training-testing splits, and the improvements of our method over the compared algorithms all passed the paired t-tests with significance value $p < 0.01$. The link prediction results are averaged across all nodes in the networks.

Firstly, by comparing the results in this section with those in Section \ref{sec:assessment}, we can clearly see that properly combining multiple meta-graphs leads to better overall performances, especially on more complicated HINs like IMDB and Yelp. Secondly, the relative performance of baselines varies across different datasets and tasks, while our method is able to constantly yield more than $10\%$ relative improvements compared with the strongest baselines on all datasets and both tasks, which clearly demonstrates its advantage. 

\begin{table*}[h!]
\centering
 \begin{tabular}{|c||c||c|c||c|c||c|}
   \hline
\multirow{2}{*}{\bf Algorithm} &\multicolumn{3}{c||}{\bf F1} & \multicolumn{3}{c|}{\bf Jaccard}\\
  \cline{2-7}
& {\bf DBLP} & {\bf IMDB} & {\bf Yelp} & {\bf DBLP} & {\bf IMDB}  & {\bf Yelp}\\
  \hline
  \hline
{\bf PTE}& $ 0.326\pm0.011 $ & $ 0.299\pm0.024 $& $ 0.031\pm0.023 $& $ 0.363\pm0.013 $& $ 0.401\pm0.039 $& $ 0.035\pm0.031 $\\
\hline
{\bf Meta2vec}& $ 0.235\pm0.012 $ & $ 0.045\pm0.008 $& $ 0.018\pm0.018 $& $ 0.262\pm0.014 $& $ 0.256\pm0.022 $& $ 0.022\pm0.031 $\\
\hline
{\bf ESim}& $ 0.596\pm0.022 $ & $ 0.290\pm0.011 $& $ 0.296\pm0.012 $& $ 0.612\pm0.022 $& $ 0.405\pm0.014 $& $ 0.244\pm0.011 $\\
\hline
{\bf HINE}& $ 0.645\pm0.013 $ & $ 0.233\pm0.007 $& $ 0.245\pm0.021 $& $ 0.648\pm0.013 $& $ 0.368\pm0.018 $& $ 0.212\pm0.021 $\\
\hline
{\bf Hin2vec}& $ 0.595\pm0.013 $ & $ 0.123\pm0.007 $& $ 0.166\pm0.015 $& $ 0.606\pm0.013 $& $ 0.287\pm0.017 $& $ 0.149\pm0.017 $\\
\hline
{\bf AspEm}& $ 0.536\pm0.014 $ & $ 0.181\pm0.028 $& $ 0.064\pm0.007 $& $ 0.591\pm0.012 $& $ 0.290\pm0.013 $& $ 0.049\pm0.011 $\\
\hline
{\bf Ours}& $ \mathbf{0.689\pm0.011} $ & $ \mathbf{0.321\pm0.021} $& $ \mathbf{0.339\pm0.015} $& $ \mathbf{0.685\pm0.019} $& $ \mathbf{0.445\pm0.013} $& $ \mathbf{0.268\pm0.012} $\\
\hline
 \end{tabular}
  \vspace{2pt}
 \caption{ \label{tab:class} Node classification performance in comparison with all baselines.}
\end{table*}

\begin{table}[h!]
\centering
\scriptsize
 \begin{tabular}{|c||c|c|c||c|c|c|}
   \hline
\multirow{2}{*}{\bf Algorithm} &\multicolumn{3}{c||}{\bf Precision@10} & \multicolumn{3}{c|}{\bf Recall@10}\\
  \cline{2-7}
& {\bf DBLP} & {\bf IMDB} & {\bf Yelp} & {\bf DBLP} & {\bf IMDB} & {\bf Yelp}\\
  \hline
  \hline
{\bf PTE}& $ 0.337 $ & $ 0.701 $& $0.212 $& $ 0.0044 $& $0.0079 $& $ 0.0023$\\
\hline
{\bf Meta2vec}& $ 0.334 $ & $ 0.623 $& $ 0.203 $& $ 0.0043 $& $ 0.0070 $& $ 0.0022 $\\
\hline
{\bf ESim}& $ 0.538  $ & $ 0.710 $& $ 0.278 $& $ 0.0070 $& $ 0.0081 $& $  0.0033$\\
\hline
{\bf HINE}& $ 0.566 $ & $ 0.727 $& $ 0.255 $& $ 0.0073 $& $ 0.0085 $& $  0.0030$\\
\hline
{\bf Hin2vec}& $ 0.548 $ & $ 0.674 $& $ 0.236 $& $ 0.0070 $& $ 0.0075 $& $ 0.0026 $\\
\hline
{\bf AspEm}& $ 0.577 $ & $ 0.663 $& $ 0.149 $& $ 0.0074 $& $ 0.0075 $& $ 0.0016 $\\
\hline
{\bf Ours}& $ \mathbf{0.612} $ & $ \mathbf{0.735} $& $ \mathbf{0.296} $& $ \mathbf{0.0079} $& $ \mathbf{0.0085} $& $ \mathbf{0.0036}  $\\
\hline
 \end{tabular}
 \vspace{2pt}
 \caption{ \label{tab:link} Link prediction performance in comparison with all baselines.}
 \vspace{-10pt}
\end{table}

\subsection{Embedding Efficiency}
\begin{SCfigure}
  \centering
  \caption{\label{fig:train}Node classification performance with varying embedding sizes and training data. The highlighted trends in our texts are even more clearly observed on IMDB and Yelp, which have more complicated network structures and larger individual embedding sizes. Due to space limit, we only present the results on DBLP. }
  \includegraphics[width=0.29\textwidth]
    {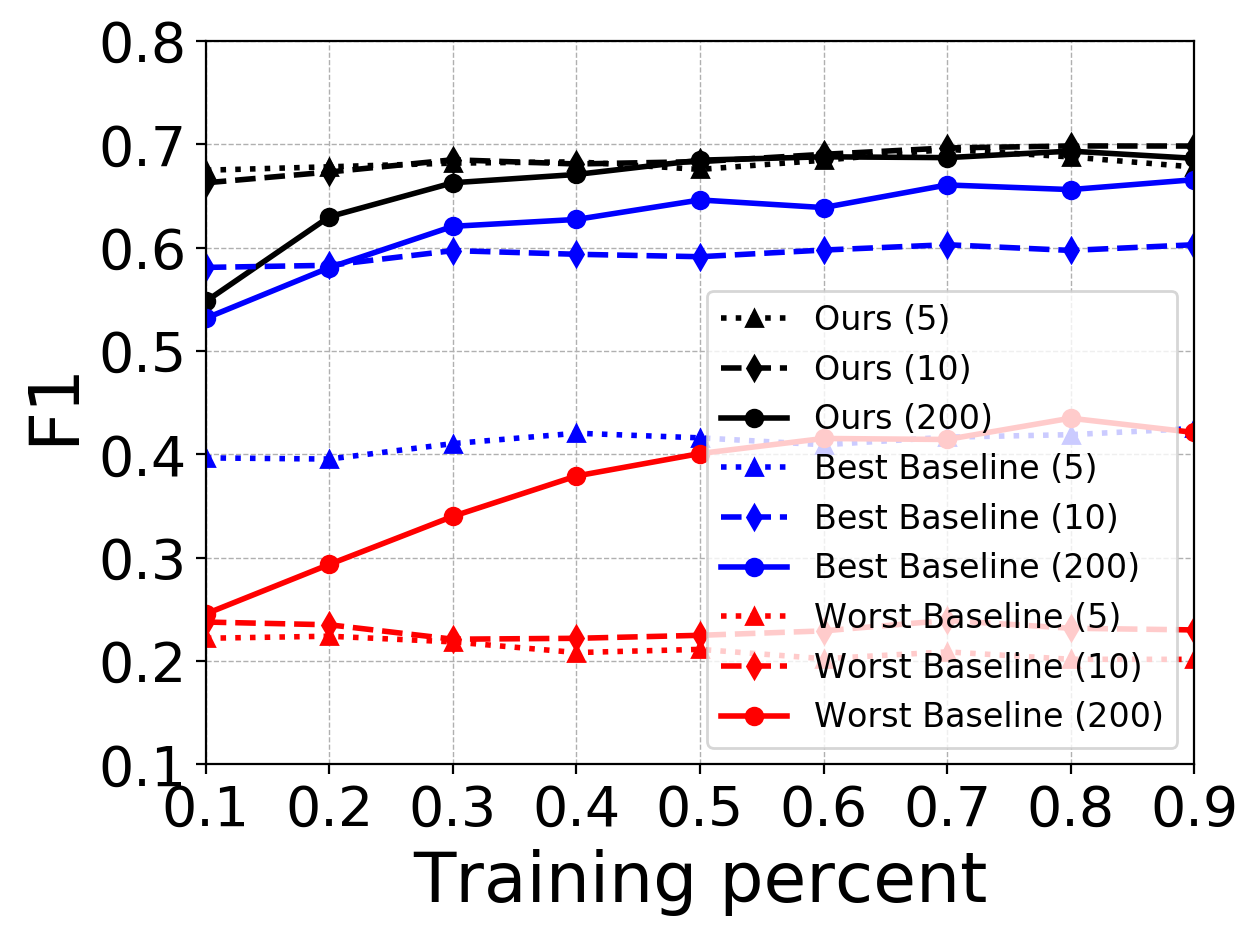}
\end{SCfigure}

Now we conduct an in-depth study on the effects of different embedding sizes and training data on the performance of our method, in order to further demonstrate our embedding efficiency.
As we can see in Figure \ref{fig:train}, for all algorithms, when the embedding size is large, the task performance relies much on the amount of training data, due to the effect of overfitting. This justifies our intuition of efficient feature selection to reduce the embedding size. On the other hand, for all baselines, small-size embeddings can hardly capture all useful information and always perform much worse than the large-size ones. Our method is the only one that efficiently captures the most important information with small embedding sizes, which is especially useful when training data are limited. 

%\begin{figure}[h!]
%\centering
%\includegraphics[width=0.3\textwidth]{figures/class_dblp.png}
%\caption{Performance with varying embedding sizes and training data.}
%\label{fig:train}
%\end{figure}

%!TEX root = hinse.tex
\section{Conclusions}
\label{sec:con}
In this work, we systematically study the assessment and combination of meta-graphs for unsupervised HIN embedding. 
%For the first part, we extend spectral graph theory to HIN settings with extensive real data analysis to arrive at a both theoretically sound and practically useful method for assessing meta-graph utility. For the second part, we devise an autoencoder with $\ell_{2,1}$-loss to simulate the process of unsupervised grouped feature selection with variance maximization. Extensive experimental results on three real-world HINs towards two traditional downstream tasks demonstrate the supreme performance of our proposed methods.
For future work, we would like to see how our methods can generally benefit various HIN models through better meta-graph selection. Moreover, our methods, while producing high-quality HIN embedding for various downstream tasks, also indicate the importance of each meta-graph in the spectral embedding process and is of great interest to in-depth studies of HIN high-order organizations in particular domains.

%!TEX root = hinse.tex
\section*{Acknowledgement}
Research was sponsored in part by U.S. Army Research Lab. under Cooperative Agreement No. W911NF-09-2-0053 (NSCTA), DARPA under Agreement No. W911NF-17-C-0099, National Science Foundation IIS 16-18481, IIS 17-04532, and IIS-17-41317, DTRA HDTRA11810026, and grant 1U54GM114838 awarded by NIGMS through funds provided by the trans-NIH Big Data to Knowledge (BD2K) initiative (www.bd2k.nih.gov).
The views and conclusions contained in this paper are those of the authors and should not be interpreted as representing any funding agencies.
%Army Research Laboratory or the U.S. Government. The U.S. Government is authorized to reproduce and distribute reprints for Government purposes notwithstanding any copyright notation hereon.

\bibliographystyle{IEEEtran}
\small
\bibliography{carlyang} 
\end{document}